\documentclass{article}

 \usepackage[preprint]{neurips_2026}

\usepackage[utf8]{inputenc} 
\usepackage[T1]{fontenc}    
\usepackage{hyperref}       
\usepackage{url}            
\usepackage{booktabs}       
\usepackage{amsfonts}       
\usepackage{nicefrac}       
\usepackage{microtype}      
\usepackage{xcolor}         
\usepackage{graphicx}
\usepackage{booktabs}
\usepackage{makecell}   
\usepackage{amssymb}    
\usepackage{multirow}
\usepackage{array}
\usepackage{tabularx}
\usepackage{makecell}
\usepackage{enumitem}
\usepackage{amsmath}
\usepackage{float}
\usepackage{marvosym}
\usepackage{subcaption}

\usepackage[table]{xcolor}
\definecolor{rowgray}{RGB}{245, 247, 250}
\definecolor{nativegreen}{RGB}{220, 240, 220}

\definecolor{headerblue}{RGB}{30, 58, 95}
\definecolor{classical}{RGB}{255, 248, 235}   
\definecolor{videovla}{RGB}{255, 240, 220}    
\definecolor{vln}{RGB}{235, 245, 255}         
\definecolor{mapbased}{RGB}{240, 255, 240}    
\definecolor{visuomotor}{RGB}{255, 235, 245}  
\definecolor{reference}{RGB}{245, 245, 245}   
\definecolor{rescuepurple}{RGB}{230, 220, 250} 
\title{RescueBench: Can Embodied Agents Save Lives \\ in the Wild?}

\author{
Kui Wu\textsuperscript{1,5} Beiyu Guo\textsuperscript{\textsuperscript{4}} Hao Chen\textsuperscript{\textsuperscript{4}} ShuHang Xu{\textsuperscript{2,5}}
Yuling Li{\textsuperscript{2,5}}
Yongdan Zeng{\textsuperscript{2,5}} \\
\bf Zhoujun Li\textsuperscript{1}
Yizhou Wang\textsuperscript{3}
Fangwei Zhong\textsuperscript{2 \Letter} \\
\textsuperscript{1}Beihang University  \textsuperscript{2}Beijing Normal University \textsuperscript{3}Peking University \\
\textsuperscript{4}City University of Macau \textsuperscript{5}ATEC2025 Challenge Committee \\ 
{{\tt\small wukui@buaa.edu.cn}, \tt \small fangweizhong@bnu.edu.cn}
}

\begin{document}

\maketitle

\vspace{-1.0cm}

\begin{figure}[H]
    \centering
    \includegraphics[width=\linewidth]{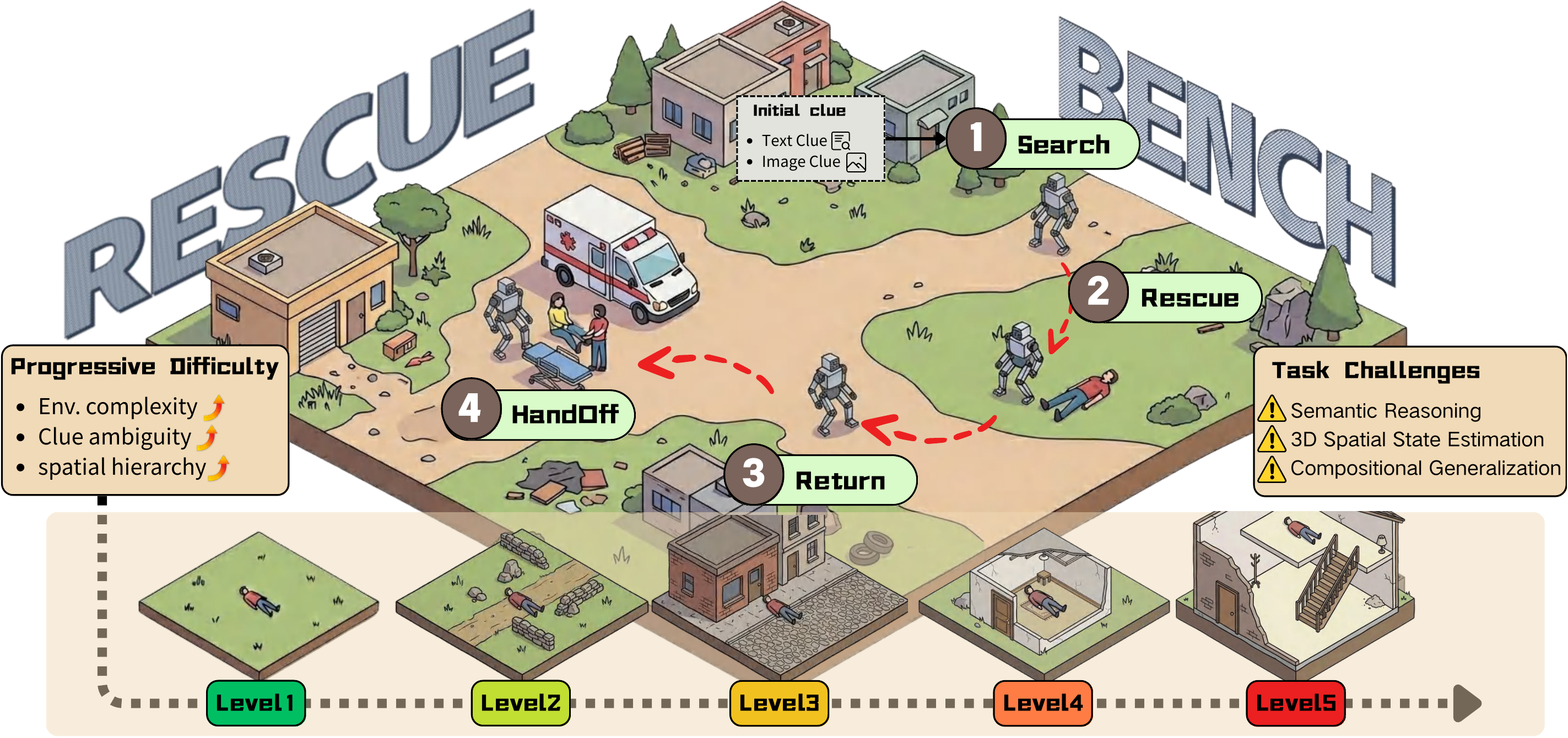}
        \vspace{-0.5cm}

    \caption{
    RescueBench comprises four sequentially dependent stages: multimodal explore (S1), locate and rescue target (S2), memory-guided return (S3), and locate and handoff (S4). This sequential composition exposes cascading failure modes absent in existing benchmarks. Five progressive difficulty levels vary along environmental complexity, clue ambiguity, and spatial hierarchy.
}
    \label{fig:task}
\end{figure}
\begin{abstract}
Search-and-rescue (SAR) requires embodied agents to explore unfamiliar environments under multimodal uncertainty, perform multi-stage interactions, and retrieve spatial memory over long horizons. Existing benchmarks typically evaluate these capabilities in isolation, leaving unclear how failures compound when they must be composed in realistic workflows. We introduce RescueBench, a photo-realistic diagnostic benchmark that instantiates SAR as a four-stage pipeline: multimodal exploration, target rescue, memory-guided return, and final handoff. By combining sequential task composition with stage-level evaluation, RescueBench enables analysis of how exploration and memory failures propagate through embodied rescue workflows. It contains five progressive difficulty levels that vary in environmental complexity, clue ambiguity, and spatial hierarchy, along with an automatic episode generation and annotation pipeline for scalable evaluation and training. We evaluate seven baselines, an oracle reference, and human players, showing that no baselines complete the full task at the greatest difficulty. Stage-level diagnosis identifies autonomous exploration as the dominant failure mode and spatial memory as a second, independent bottleneck, suggesting that these limitations are not resolved by current topological visual-language navigation or map-based methods. Code is available in \url{https://github.com/UnrealZoo/RescueBench}.

\end{abstract}

\section{Introduction}
\label{sec:intro}
Search-and-rescue (SAR) represents a critical application scenario in embodied intelligence~\cite{QUERO2025105199,zhou2024hazard}, where agents must operate autonomously in unstructured, hazardous environments to locate and extract victims. At its core, SAR demands three integrated capabilities: autonomous exploration under multimodal uncertainty, multi-stage physical interaction, and spatial memory retrieval over large-scale environments. Yet existing benchmarks evaluate these capabilities in isolation, e.g., vision-language navigation (VLN) benchmarks evaluate step-wise instruction-following \cite{anderson2018vision,krantz2020beyond,wang2023learning}, manipulation benchmarks test fine-grained tabletop interaction \cite{liu2023libero,mees2022calvin}, and outdoor navigation benchmarks test goal-reaching with known destinations \cite{liu2025citywalker,mei2026urbannav, wang2025trackvla, liu2025trackvla++, lin2025embrace3kembodiedreasoningaction}. No benchmark unifies open-ended exploration, sequential physical interaction, and spatial memory under progressive difficulty, nor exposes how failures cascade when these capabilities must be composed in sequence. To bridge this gap, we introduce RescueBench, a large-scale photo-realistic benchmark designed to capture the operational complexity and cascading decision demands that real-world SAR entails.

Recent benchmarks have shifted toward evaluating general-purpose multimodal large language models (MLLMs) through prompt engineering \cite{yang2025embodiedbench, zhong2025unrealzoo, puig2023habitat, ren2025simworld, wu2025vlm}, which does not represent the specialized embodied architectures developed by the community. Existing embodied benchmarks~\cite{mei2026urbannav,liu2023libero,shridhar2020alfred, qiu2017unrealcv, wu2025hierarchicalinstructionawareembodiedvisual, wu2026adatracker} remain largely confined to single-task evaluation, either navigation or object manipulation, in structured indoor environments~\cite{puig2023habitat,anderson2018vision,krantz2020beyond}, while data collection infrastructure is limited to navigation trajectories without interaction annotations ~\cite{ci2023proactive, wang2025communication}. This narrow scope falls short of the continuous perception-action loops, open-world exploration, and physical interaction demands that outdoor SAR entails.

\begin{table}[!b]
\centering
\caption{
Comparison of SAR benchmarks across five evaluation dimensions. 
\textbf{Guidance}: step-by-step instruction, waypoints, or multimodal clues 
requiring active search. \textbf{Task}: single-stage, multi-stage, or sequential (seq.). 
\textbf{Multi-level Diff}: number of progressive levels. 
\textbf{Cooperation}: multi-agent cooperation support. See footnotes for details.
}
\label{tab:benchmark_comparison}
\resizebox{\textwidth}{!}{%
\begin{tabular}{l c c c c c c}
\toprule
\rowcolor{headerblue}
\textcolor{white}{\textbf{Benchmark}} &
\textcolor{white}{\textbf{Env.}} &
\textcolor{white}{\textbf{Guidance}} &
\textcolor{white}{\textbf{Task}} &
\textcolor{white}{\textbf{Interaction}} &
\textcolor{white}{\textbf{Multi-level Diff}} &
\textcolor{white}{\textbf{Cooperation}} \\
\midrule
\rowcolor{vln}
R2R \cite{anderson2018vision}
  & Indoor & Instruction & Single & $\times$ & 1 & $\times$ \\
\rowcolor{vln}
VLN-CE \cite{krantz2020beyond}
  & Indoor & Instruction & Single & $\times$ & 1 & $\times$ \\
\rowcolor{vln}
EVT \cite{zhong2024empowering}
  & Indoor+Outdoor & Instruction & Single & $\times$ & 1 & $\times$ \\
\rowcolor{classical}
ALFRED \cite{shridhar2020alfred}
  & Indoor & Instruction & Multi & $\checkmark$ & 1 & $\times$ \\
\rowcolor{classical}
Habitat~3.0 \cite{puig2023habitat}
  & Indoor & Instruction & Multi & $\checkmark$ & 1 & $\times$ \\
\rowcolor{classical}
LIBERO \cite{liu2023libero}
  & Table & N/A & Multi  & $\checkmark$ & 1 & $\times$ \\
\rowcolor{classical}
CALVIN \cite{mees2022calvin}
  & Table & N/A & Multi & $\checkmark$ & 1 & $\times$ \\
\rowcolor{mapbased}
CityWalker \cite{liu2025citywalker}
  & Outdoor & Waypoints & Single & $\times$ & 1 & $\times$ \\
\rowcolor{mapbased}
UrbanNav \cite{mei2026urbannav}
  & Outdoor & Instruction & Single & $\times$ &1 & $\times$ \\
\rowcolor{visuomotor}
PARTNR \cite{changpartnr}
  & Indoor & Instruction & Multi & $\checkmark$ & 1 & $\checkmark$ \\
\rowcolor{visuomotor}
COHERENT \cite{liu2025coherent}
  & Indoor & Instruction & Multi & $\checkmark$ & 1& $\checkmark$ \\
\midrule
\rowcolor{rescuepurple}\textcolor{black}{\textbf{RescueBench}} &\textcolor{black}{\textbf{Indoor+Outdoor}} &\textcolor{black}{\textbf{Multimodal clue}} &\textcolor{black}{\textbf{Multi (Seq.)}} &\textcolor{black}{$\boldsymbol{\checkmark}$} &\textcolor{black}{\textbf{5}} &\textcolor{black}{$\checkmark$\textsuperscript{$\dagger$}} \\
\bottomrule
\end{tabular}%
}
{\footnotesize
\quad
$\dagger$~The RescueBench task framework supports heterogeneous multi-agent 
configurations; this submission focuses on single-agent evaluation, with 
preliminary multi-agent validation provided in Appendix~\ref{app:multi_agent}.
}
\end{table}

We introduce RescueBench, which comprises: (1)~\textbf{A cascading-failure SAR benchmark} that composes exploration, physical interaction, and spatial memory into a single four-stage pipeline (explore, locate-and-rescue, return, locate-and-handoff), enabling stage-level diagnosis of where and how embodied agents fail under sequential task demands. (2)~\textbf{A progressive difficulty framework} with five levels that independently vary environmental complexity, clue ambiguity, and spatial hierarchy, allowing controlled analysis of how specific environmental factors affect each capability. (3)~\textbf{An empirical diagnosis of exploration and spatial memory as persistent bottlenecks} across seven methods spanning four architectural families, revealing that open-world exploration is absent in current VLN methods and that spatial memory deficits persist even after successful target localization.

Evaluation of seven learned baselines, an oracle reference, and a human performance reveals two principal findings. First, no existing architecture completes the full task at the highest difficulty level ($L_5$: cross-region indoor--outdoor navigation with multi-floor structural complexity): autonomous exploration under uncertainty is the dominant failure mode, and open-world exploration is not an emergent capability of topological graph-based VLN or map-based paradigms but a persistent architectural limitation in the evaluated families. Second, spatial memory retrieval, i.e., navigating back to a known location without step-wise guidance, emerges as an independent bottleneck: even when agents successfully locate the target, they frequently fail to return to the ambulance, revealing that the capability deficit extends beyond exploration. Unlike benchmarks that measure each capability in isolation, RescueBench measures the composite capability, whether an agent can sustain exploration, interaction, and memory retrieval in sequence, which is the prerequisite for any multi-stage SAR deployment. RescueBench thus provides the diagnostic infrastructure to pinpoint where and why embodied agents fail, exposing capability gaps that only emerge under sequential task composition.

\section{RescueBench}
RescueBench is a photo-realistic embodied AI benchmark designed to evaluate whether current agents can execute the full search-and-rescue pipeline in unstructured environments. We introduce RescueBench through three components: (i) a four-stage sequential task definition that mirrors real-world SAR protocols and isolates exploration, approach navigation, and spatial memory capabilities; (ii) five progressive difficulty levels that systematically vary environmental complexity, clue ambiguity, and spatial hierarchy; and (iii) a scalable automatic data collection pipeline built on Unreal Engine 5. The design intentionally decouples evaluation from manipulation limitations through environment-assisted interaction triggering, ensuring that failure diagnosis reflects architectural gaps in navigation and exploration rather than manipulation primitives. The following sections detail each component.
\begin{figure}[tb]
    \centering
    \includegraphics[width=0.95\linewidth]{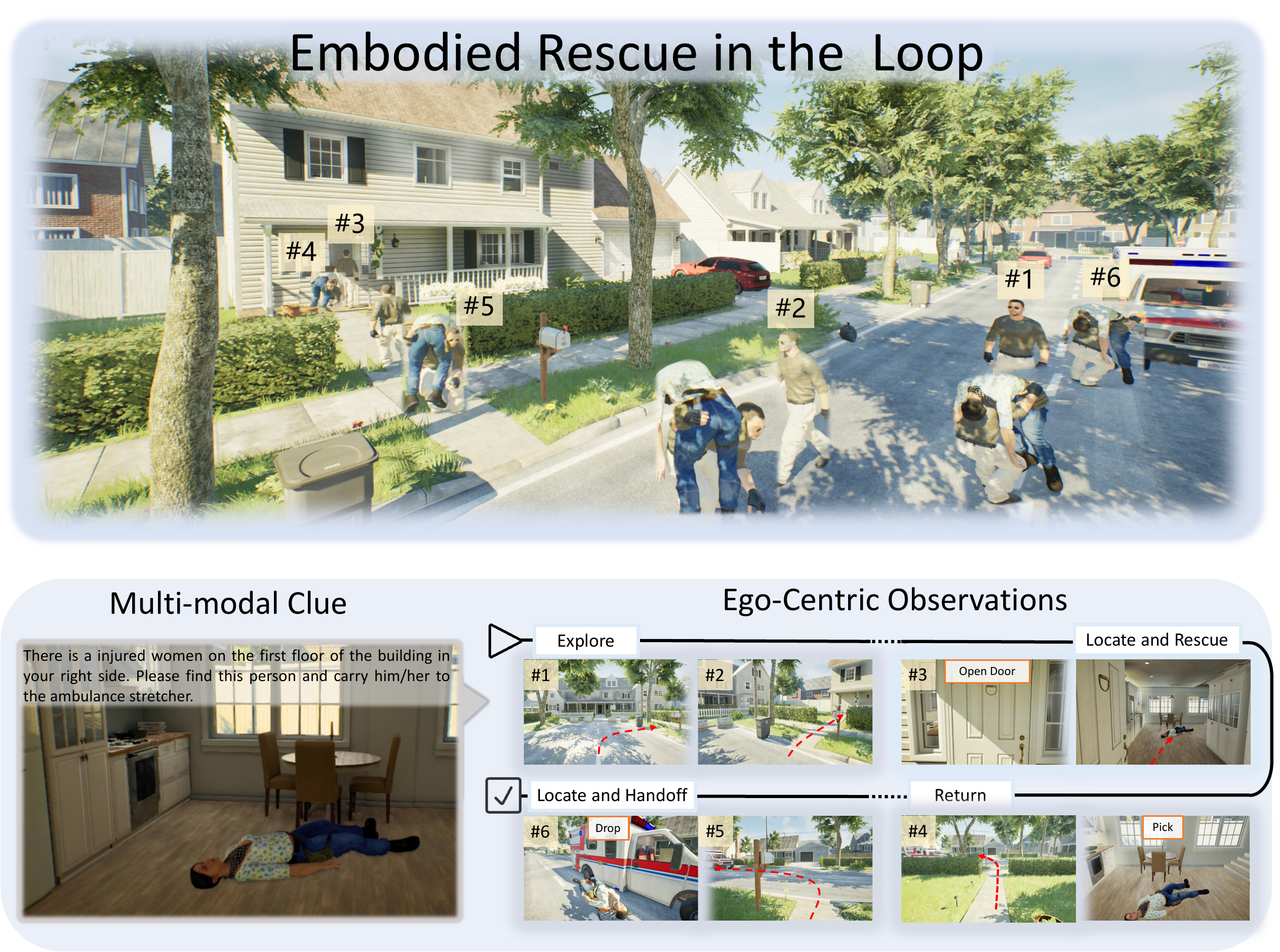}
\caption{A complete $L_4$ episode: the agent explores from an outdoor area, opens a door to traverse indoors (bottom inset) (S1), locates the injured person (S2), returns through the same path (S3), and completes handoff at the stretcher zone (S4).}
    \label{fig:task_demo}
    \vspace{-0.5cm}
\end{figure}

\subsection{Task Definition}
We formulate search-and-rescue as a four-stage sequential task that mirrors the real-world SAR pipeline. An ambulance with a stretcher (home base) is positioned at $\mathbf{p}_{\text{amb}} \in \mathcal{E}$, and an injured target at $\mathbf{p}_{\text{tar}} \in \mathcal{E}$. A single agent, acting as the primary rescuer, is tasked with all physical interactions. The formulation naturally generalizes to heterogeneous multi-agent configurations (Appendix~\ref{app:multi_agent}), but we currently focus on the single-agent setting for fine-grained evaluation. The task requires agents to operate through continuous perception-action loops: observing visual scenes, estimating spatial relationships, and executing continuous motion commands in a photo-realistic 3D world.


Each stage isolates a distinct embodied capability:
\textbf{Stage 1 — Explore.} Given only sparse multimodal clues $\mathcal{C}_{\text{multi}} = {V_{\text{sparse}}, T_{\text{text}}}$ (e.g., bottom left in Figure~\ref{fig:task_demo}) with no navigation instructions or target coordinates, the agent must autonomously formulate an exploration strategy to discover the target within the detection range. \textbf{Stage 2 — Locate and Rescue.} The agent navigates to the discovered target and triggers a rescue interaction within a suitable proximity range. For agents lacking native manipulation capability, we employ environment-assisted triggering once the agent enters the interaction proximity zone. This design choice ensures that evaluation is not bottlenecked by manipulation limitations for navigation-focused architectures, while still testing their core competency, namely reaching the correct location under uncertainty. \textbf{Stage 3 — Return.} The agent navigates back to the ambulance at $\mathbf{p}_{\text{amb}}$ by retrieving its location from spatial memory accumulated during S1, testing in-context navigation without step-wise goal instructions. \textbf{Stage 4 — Locate and Handoff.} The agent precisely locates the ambulance within the scene and triggers a placement interaction, evaluating precise approach navigation to a known spatial target. Full task completion requires $\text{S1} \wedge \text{S2} \wedge \text{S3} \wedge \text{S4}$. Each stage can also be evaluated independently, enabling fine-grained diagnosis of capability bottlenecks. 

\textbf{Design Rationale and Scope.}
Three design choices shape the benchmark's diagnostic target. First, S1 measures exploration-under-uncertainty via multimodal clue reasoning rather than pure instruction following, reflecting the information regime of real SAR where rescuers interpret incomplete cues. Second, S3 probes return-by-memory; the diagnostic (S1,S2)--S3 gap isolates the quality of spatial encoding accumulated during the S1 exploration phase, rather than local navigation ability alone. Third, environment-assisted interaction triggering abstracts away manipulation to isolate navigation, search, and memory bottlenecks, broadening applicability to agents without manipulation primitives.

\subsection{Progressive Difficulty Levels}
\begin{figure}[!tb]
    \centering
    \includegraphics[width=\linewidth]{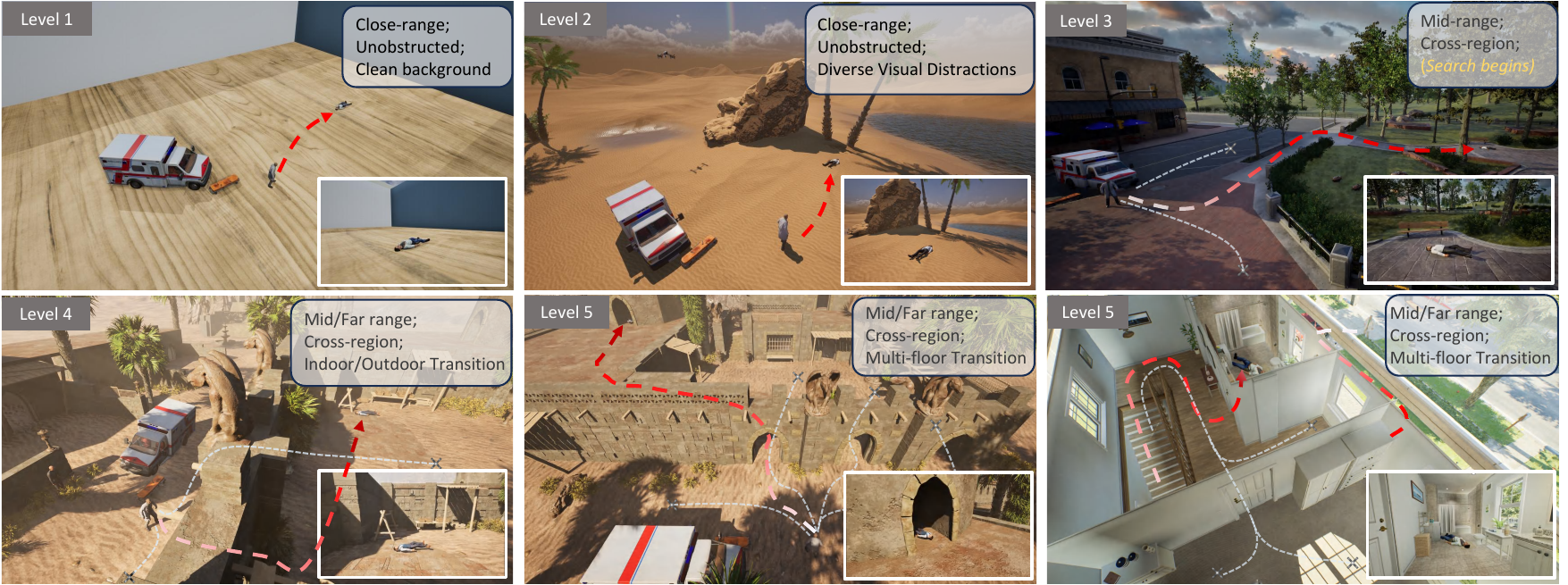}
\caption{Progressive difficulty levels $L_1$--$L_5$ across test environments. Each level escalates spatial complexity and environmental diversity: $L_1$--$L_2$ place targets at close range in clean or cluttered settings, while $L_3$--$L_5$ require sustained search across larger spatial hierarchies. The white dashed lines illustrate that high-difficulty episodes demand agents to actively explore and select among multiple feasible routes, a decision-making capability absent in route-following paradigms.}

    \label{fig:progress_diff}
\end{figure}

At every difficulty level, the agent must complete the full four-stage task. The five difficulty levels $\mathcal{L} = \{L_1, \ldots, L_5\}$ control task difficulty along three axes: environmental complexity, clue ambiguity, and spatial hierarchy (Figure~\ref{fig:progress_diff}). The $L_2 \!\to\! L_3$ transition, from close-range visible targets to mid-range occluded targets requiring active search, produces the steepest capability cliff across all evaluated methods. $L_3 \!\to\! L_5$ progressively compound exploration demands with cross-region wayfinding ($L_4$) and multi-floor structural navigation ($L_5$). In parallel, clue ambiguity escalates from visual matching ($L_3$) through directional reasoning ($L_4$) to hierarchical spatial decomposition ($L_5$), mirroring real-world SAR information regimes; the corresponding natural-language reference instructions also vary in length and complexity across levels (statistics in Figure~\ref{fig:testset_overview}). 
Target placement at each level is governed by a level-specific distance range combined with two validation filters. Details are provided in Appendix~\ref{app:difficulty}.

\subsection{Environments and Simulator}
RescueBench is built on UnrealZoo~\cite{zhong2025unrealzoo}, a UE5-based platform providing photo-realistic environments with built-in rendering control via UnrealCV~\cite{qiu2017unrealcv}. From this pool, we select 7 representative scenes spanning residential, commercial, outdoor street, and natural terrain, with layouts ranging from single-floor flat to multi-level structures. All rescue-specific components (task execution logic, interaction systems, SAR action mappings) are built from scratch. The platform accepts any UnrealZoo environment with a valid navigation mesh, making the full scene pool available for evaluation expansion without platform modification. Details are provided in Appendix~\ref{app:simulator}.

\begin{figure}
    \centering
    \includegraphics[width=\linewidth]{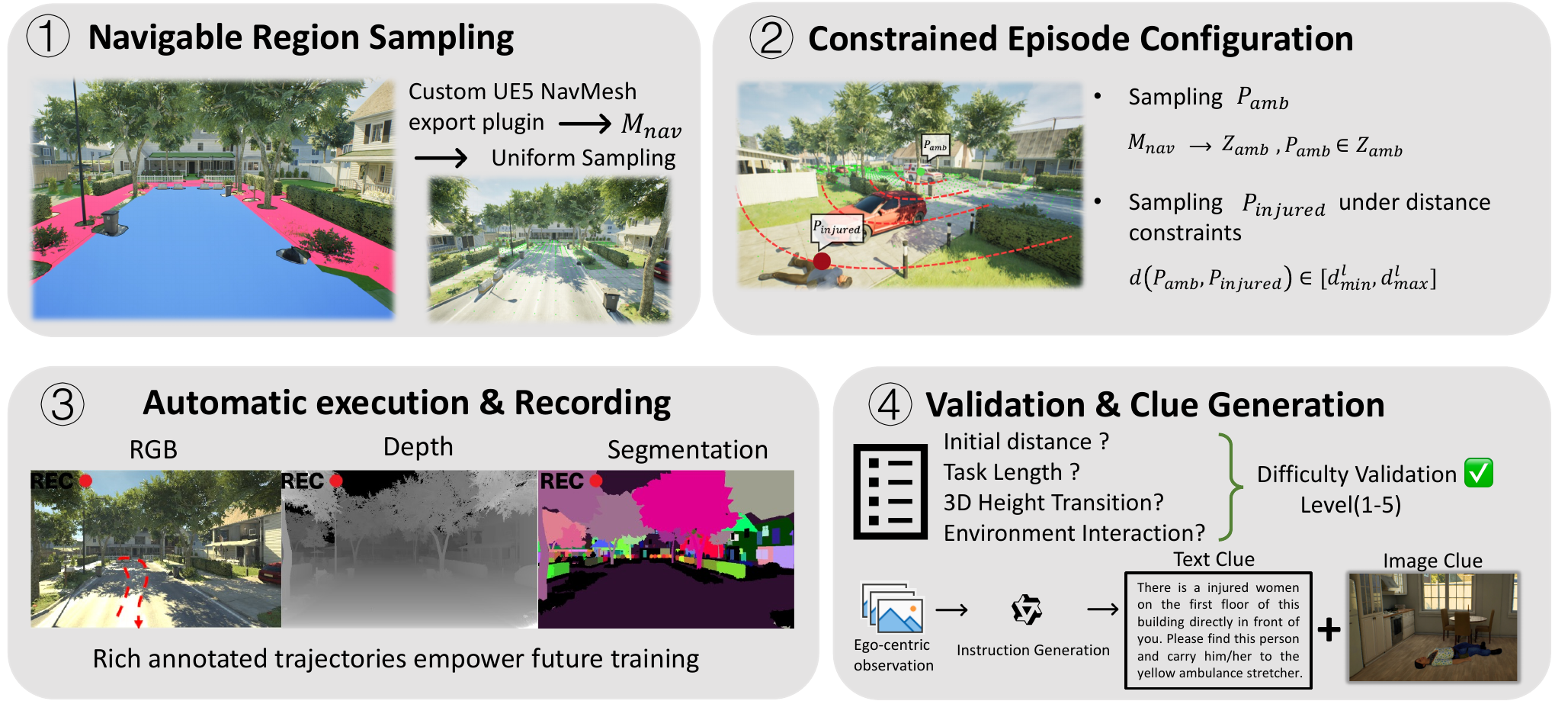}
    \caption{Scalable data collection pipeline. Given environment assets and a UE5 navigation mesh, the pipeline automates episode generation and annotation in four steps: (1) candidate spawn zones are extracted from the navigation mesh; (2) ambulance, target, and agent positions are sampled under level-specific constraints; (3) a navmesh-driven agent executes the task while trajectories and interaction annotations are recorded; (4) validation rules enforce difficulty consistency, and multimodal clues are generated from the agent's egocentric observations via a VLM (Qwen-3.5 plus).
}
    \label{fig:data_collection}
    \vspace{-0.5cm}
\end{figure}

\subsection{Scalable Data Collection Pipeline}

Existing automatic pipelines capture only navigation trajectories without interaction annotations~\cite{wang2025trackvla, liu2025trackvla++,zhong2024empowering}. RescueBench introduces a scalable data collection pipeline (Figure~\ref{fig:data_collection}) that automates episode generation and annotation once environment assets and a valid UE5 navigation mesh are provided. Given this setup, the pipeline samples candidate positions from the navigation mesh, configures episodes under difficulty constraints, drives an internal navigation agent through the full four-stage SAR workflow while recording the complete trajectory with interaction annotations (pre- and post-interaction observations, waypoint sequences, RGB/depth/segmentation snapshots), and validates each episode against its intended difficulty label. Unlike navigation-only pipelines, this records the complete interaction stream, reducing manual annotation to zero per episode while providing dense supervision for training approach behaviors. Clues for $L_3$--$L_5$ are auto-generated from the expert's first-person observation sequence using a VLM. Details are listed in Appendix~\ref{app:pipeline}.

\section{Experiments}
\label{sec:exp}

We evaluate seven learned baselines, one oracle reference, and human performance to answer three questions: (RQ1) how close are current methods to solving the full four-stage SAR pipeline and where do they plateau; (RQ2) which specific capability (exploration, precise approach, or spatial memory) accounts for the majority of failures; and (RQ3) what do these failure patterns reveal about architectural limitations across paradigm families. 

\textbf{Evaluation Protocol.}
All baselines operate under a unified protocol. \emph{Observations:} every method receives synchronized RGB, depth, and semantic segmentation at each step; each method uses its official default input resolution (listed in Appendix~\ref{app:simulator}). \emph{Action interface:} ground agents share a continuous action space (forward velocity, angular velocity) with three binary interaction actions (pick, drop, open door); method-specific wrappers translate native outputs into this interface. \emph{Interaction mechanism:} all methods except ROCKET-2 use environment-assisted triggering ($\delta_{\text{trig}} = 1.5$\,m proximity), while ROCKET-2 uses its native manipulation actions; both trigger identical task state transitions. \emph{Resource budgets:} episodes are governed by level-dependent time limits ($L_1$--$L_2$: 180\,s, $L_3$: 240\,s, $L_4$--$L_5$: 300\,s) with no step cap; the run-time fps is determined by each method's inference latency rather than the environment. \emph{Fine-tuning:} the four adapted methods (Uni-NaVid, ViNT, NoMaD, ROCKET-2) are fine-tuned on the same automatically collected dataset ($\approx$400K steps, 9:1 train/validation split); all other methods are evaluated zero-shot with their original pretrained weights.  Full protocol details and fine-tuning hyperparameters are provided in Appendix~\ref{app:simulator} and ~\ref{app:finetune}.


\subsection{Baselines}

We evaluate seven learned baselines spanning five architectural families: classical two-stage pipelines, video VLA, VLN foundation models, map-based method, and goal-conditioned visuomotor policies. We also include two non-learning references (Table~\ref{tab:baselines}). The references: human players and the Oracle Navigator, contextualize performance against biological capability and the theoretical upper bound. The framework also supports multi-agent evaluation. Preliminary multi-agent experiments and detailed method descriptions are provided in Appendix~\ref{app:multi_agent} and~\ref{app:baseline_details}. 

\begin{table}[t]
\centering
\caption{Evaluated methods. \textbf{Type}: architectural paradigm. \textbf{Core Mechanism}: how navigation decisions are produced. \textbf{Interaction}: whether the architecture natively supports physical interaction or requires the environment-assisted triggering mechanism. \textbf{Fine-tuned}: methods fine-tuned on RescueBench expert demonstrations (Section~\ref{sec:exp}).}
\label{tab:baselines}
\resizebox{\textwidth}{!}{%
\begin{tabular}{l l l c c}
\toprule
\rowcolor{headerblue}
\textcolor{white}{\textbf{Method}} & 
\textcolor{white}{\textbf{Type}} & 
\textcolor{white}{\textbf{Core Mechanism}} & 
\textcolor{white}{\textbf{Interaction}} & 
\textcolor{white}{\textbf{Fine-tuned}} \\
\midrule
\rowcolor{classical}
LLM-YOLO Planner~\cite{Jocher_Ultralytics_YOLO_2023} & Classical Two-Stage & Detection + GPT-4o planning & Env-assisted & --- \\
\rowcolor{videovla}
Uni-NaVid \cite{zhang2024uni} & Video VLA & Video-conditioned action prediction & Env-assisted & \checkmark \\
\rowcolor{vln}
ViNT \cite{shah2023vint} & VLN Foundation & Topological graph waypoint retrieval & Env-assisted & \checkmark \\
\rowcolor{vln}
NoMaD \cite{sridhar2024nomad} & VLN Foundation & Goal-masking diffusion policy & Env-assisted & \checkmark \\
\rowcolor{mapbased}
SG-Nav \cite{yin2024sg} & Map-Based & Online 3D scene graph + LLM & Env-assisted & --- \\
\rowcolor{mapbased}
OmniNav \cite{xue2026omninav} & Map-Based & Vision-language grounding on map & Env-assisted & --- \\
\rowcolor{visuomotor}
ROCKET-2 \cite{cai2025rocket} & Visuomotor Policy & Cross-view goal alignment & \textbf{Native} & \checkmark \\
\midrule
\rowcolor{reference}
Human Player & Reference & Keyboard/mouse control & \textbf{Native} & --- \\
\rowcolor{reference}
Oracle Navigator & Reference & Ground-truth path + state machine & \textbf{Native} & --- \\
\bottomrule
\end{tabular}%
}
\end{table}

\subsection{Evaluation Metrics}
\label{subsec:metrics}

\textbf{Task Completion Rate (TCR)} Binary success indicators decompose the sequential pipeline: 

\begin{equation}
    \text{TCR} = \frac{|\{i : \text{S1}_i \wedge \text{S2}_i \wedge \text{S3}_i \wedge \text{S4}_i\}|}{N},
\end{equation}
\textbf{Task Score (TS).}
A continuous $[0, 100]$ score measuring partial progress, awarding credit proportional to proximity achieved at each stage:
\begin{equation}
  \text{TS}_i = \sum_{k \in \{1,2,3,4\}} \text{StageScore}_{k,i}, \quad
  \text{StageScore}_{k,i} = 25 \cdot \text{clip}\!\left(1 - \frac{d_k^{\text{best}}}{\max(d_k^{\text{init}}, \epsilon)},\; 0,\; 1\right).
\end{equation}
TS provides a dense optimization signal for training and enables ranking of methods that achieve similar TCR but differ in near-miss behavior.


\textbf{Efficiency: Average Time Cost and Avg Steps.} Two complementary metrics capture the computational cost and behavioral efficiency of each method. \textbf{Average Time Cost} reports the mean episode wall-clock duration in seconds, reflecting the end-to-end deployment cost. \textbf{Avg Steps} counts the number of inference steps the agent executes per episode. Together, they distinguish two sources of inefficiency: high per-step latency (few steps but long wall-clock time) versus inefficient exploration (many steps but little progress).

\textbf{Human Similarity} To capture trajectory-level strategy rather than step-level precision, we measure alignment between agent and human trajectories using Dynamic Time Warping (DTW)~\cite{sakoe1978dynamic}, normalized by median human path length. HS values are reported in the appendix tables; the full definition and diagnostic interpretation are provided in Appendix~\ref{app:hs_detail}.


Each metric category provides a distinct diagnostic lens: TCR and TS identify \emph{which} pipeline stage fails and \emph{how close} the agent came to success; Average Time Cost and Avg Steps reveal \emph{how efficiently} the agent explores; and HS quantifies \emph{how similarly} the agent behaves compared to human capability. Together, they enable multi-dimensional capability profiling across exploration under uncertainty, spatial reasoning for interaction timing, and in-context navigation via spatial memory.

\begin{figure}
\vspace{-0.5cm}
    \centering
    \includegraphics[width=\linewidth]{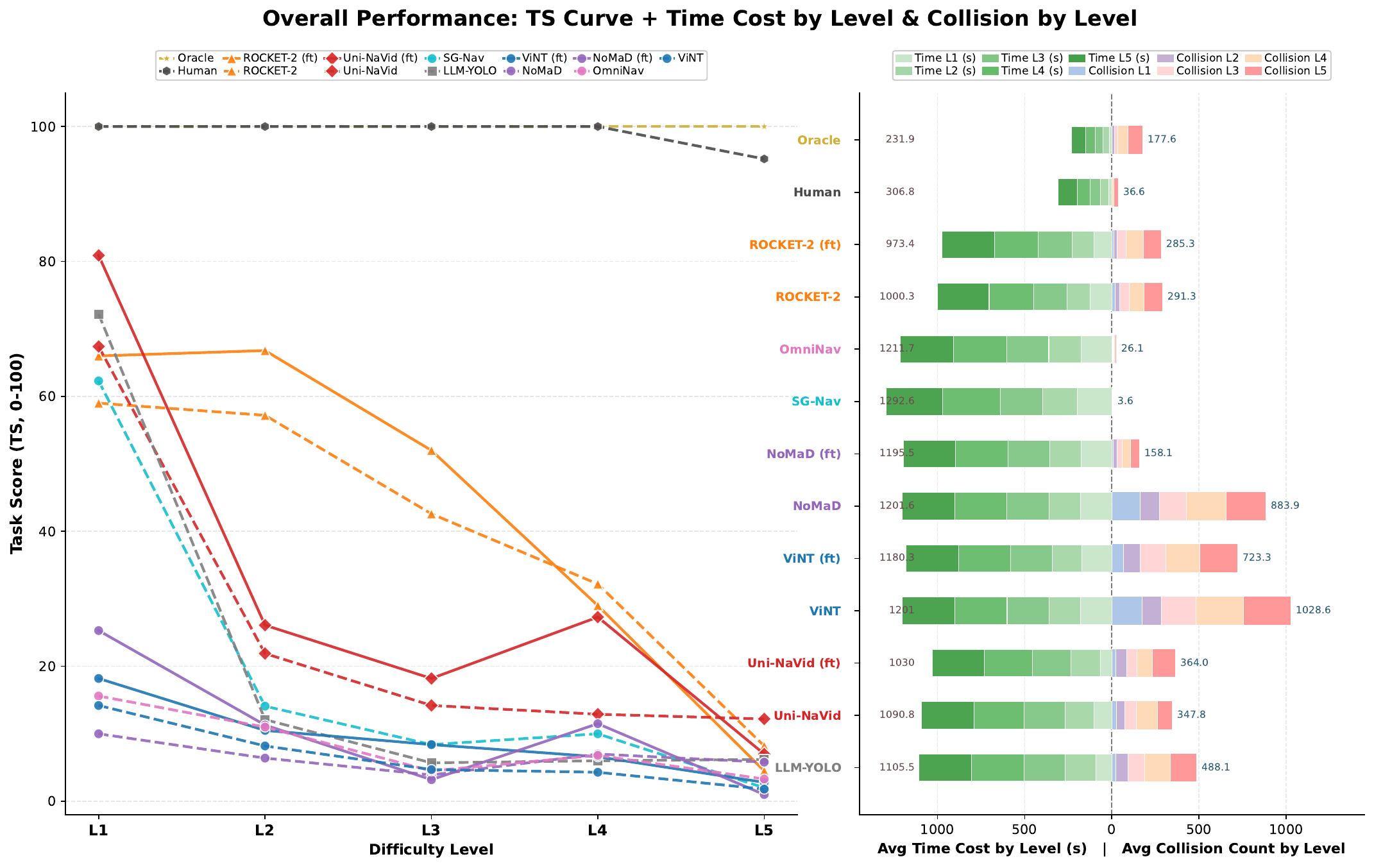}
    \caption{
    Task Score degradation across five difficulty levels (left) for both zero-shot and fine-tuned version (followed by ``(ft)''), and average time cost vs.\ average collision count (right) per method. The capability cliff at
$L_1$→$L_2$ reflects collapse under visual complexity. ROCKET-2(ft) achieves the highest TS at $L_3$ (52.0). No method achieves non-zero full-task completion at $L_5$. The right panels expose three behavioral regimes: inference-limited (map-based: few steps, time-saturated), exploration-competent (VLA: moderate cost, systematic coverage), and movement-without-exploration (VLN: high cost, degenerate looping). Detailed metric values are listed in Appendix.
}
    \label{fig:zero_shot_benchmark}
    \vspace{-0.5cm}
\end{figure}
\subsection{Overall Performance}
Figure~\ref{fig:zero_shot_benchmark} summarizes performance across all five difficulty levels. No existing paradigm completes the full four-stage pipeline at $L_5$; detailed metric values for all methods are listed in the appendix.

\paragraph{Zero-shot Capability Landscape.}

Under zero-shot evaluation, this gap unfolds progressively in Task Score. The $L_1 \to L_2$ transition produces a sharp capability cliff: LLM-YOLO Planner's TS drops from 72.2 to 12.1 as visual diversity disrupts its modular detection pipeline. From $L_2 \to L_3$, the shift from close-range targets to active search degrades all methods further, with ROCKET-2 achieving the highest zero-shot $L_3$ TS (42.6). Beyond $L_3$, spatial hierarchy (cross-region at $L_4$, multi-floor at $L_5$) collapses all methods to near-zero Task Score. Navigation foundation models (ViNT, NoMaD) and map-based approaches (SG-Nav, OmniNav) consistently perform poorly across all levels, with ViNT and NoMaD failing to achieve any full-task completion even at $L_1$. 

The right panels of Figure~\ref{fig:zero_shot_benchmark} further expose three behavioral regimes along the time dimension: map-based methods saturate the time budget despite executing few steps (inference-limited); VLN models accumulate high collision counts with degenerate looping that fails to explore new areas (movement without exploration); and VLA methods achieve sufficient spatial coverage for partial task progress within the budget (exploration-competent).

Crucially, these regimes indicating that time efficiency is governed by exploration strategy rather than inference frequency, and strategy is in turn dictated by architecture. In real-world SAR deployment, where every second of response time directly affects survival outcomes, this efficiency gap between paradigms is not merely an evaluation artifact but a practical constraint on deployability.

\paragraph{Fine-tuning Diagnostic Gains.}
To validate whether the automatic data collection pipeline provides effective supervision, we fine-tune four trainable methods (Uni-NaVid, ViNT, NoMaD, ROCKET-2; details in Appendix~\ref{app:finetune}). Fine-tuning yields substantial TCR gains: Uni-NaVid doubles the completion rates at $L_1$ (24\%$\to$52\%), while VLN models (ViNT and NoMaD) break through from 0\% to 8\%. ROCKET-2 (fine-tuned) achieves the highest overall $L_3$ TS (52.0). However, these gains plateau sharply beyond $L_1$: even after adaptation, VLN models only reach 8\% TCR at the simplest level, and all fine-tuned methods collapse to near-zero at $L_4$ and $L_5$. This architecture-dependent ceiling confirms that the bottleneck is not data scarcity but the absence of exploration logic and spatial memory in paradigms designed for route retrieval.


    \vspace{-0.5cm}
\subsection{Stage-Level Diagnosis}
\label{subsec:stage_diagnosis}
\begin{figure}
    \vspace{-0.5cm}

    \centering
    \includegraphics[width=\linewidth]{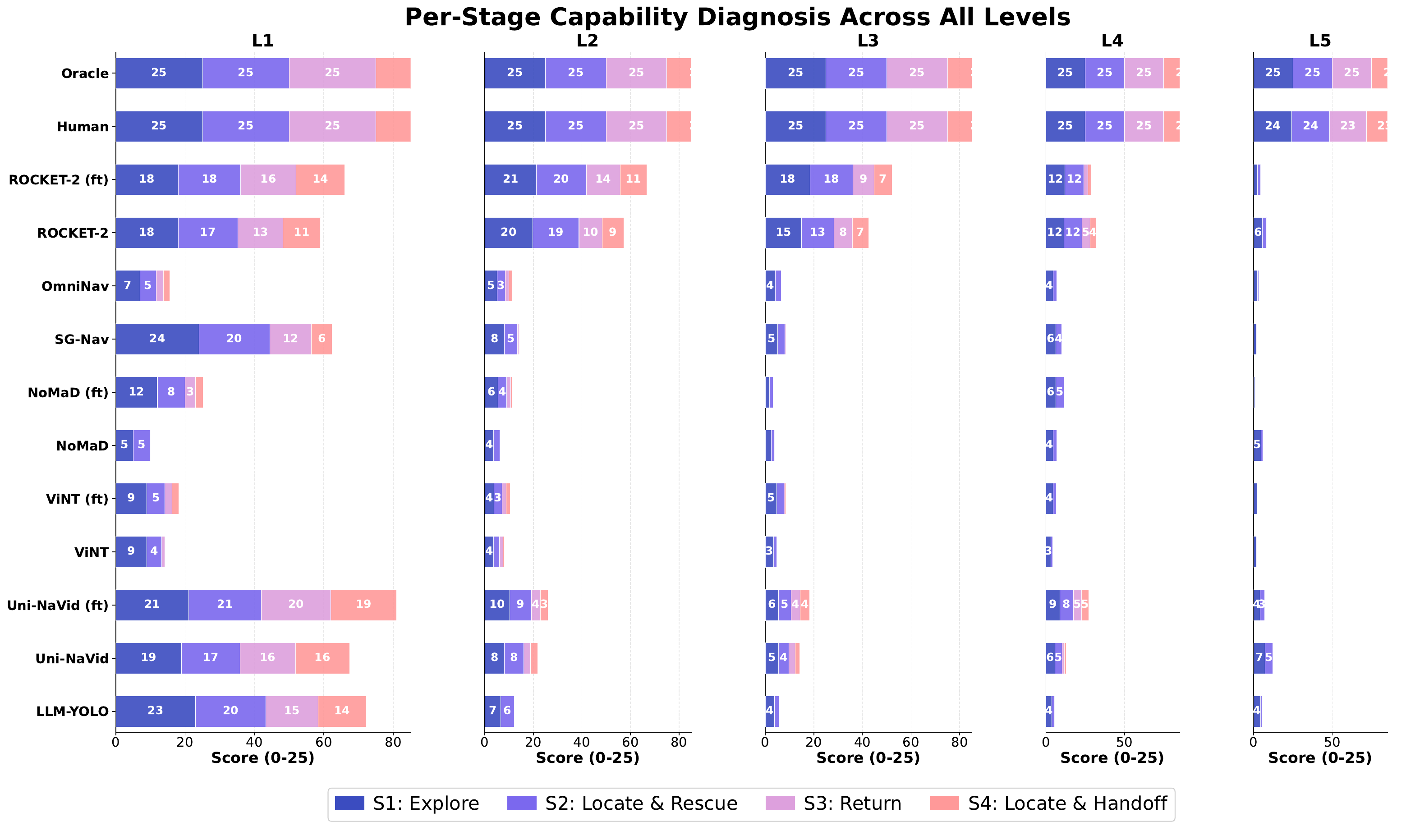}
    \vspace{-0.5cm}

    \caption{Per-stage Task Score (/25) across five difficulty levels, decomposed by method. Each stacked bar separates S1 (Explore), S2 (Locate and Rescue), S3 (Return), and S4 (Locate and Handoff), revealing two principal bottlenecks: exploration collapse under uncertainty (S1) and spatial memory degradation for compositional return navigation (S3).}
    \label{fig:per_stage_analysis}
    \vspace{-0.5cm}

\end{figure}

To diagnosis the agents at different stage, we decomposes each method's TS into per-stage contributions, revealing two principal bottlenecks, as shown on Figure~\ref{fig:per_stage_analysis}. The detailed per-stage values are reported in Appendix Table~\ref{tab:stage_all_1} and~\ref{tab:stage_all_2}.

\textbf{S1 (Exploration under multimodal uncertainty)} is the dominant bottleneck. In Figure~\ref{fig:per_stage_analysis}, most methods remain near the bottom of the S1 range from $L_3$ onward. Even the strongest exploration result (ROCKET-2, fine-tuned: 18.4/25 at $L_3$) drops sharply at higher levels (12.2 at $L_4$, 2.4 at $L_5$), indicating that exploration degrades once search must be sustained across larger spatial hierarchies.

\textbf{S2 (Locate \& Rescue)} and \textbf{S4 (Locate \& Handoff)} track S1 closely across all methods, with gaps of only 0--3 points. This implies that once a target is located, current models navigate to it and carry out the required interaction with reasonable reliability. The primary challenge is in finding the target, not in approaching it.

\textbf{S3 (Spatial memory for compositional navigation)} introduces a second, independent bottleneck. Whenever non-zero return-stage performance is achieved, S3 is visibly weaker than S2. For example, ROCKET-2 (fine-tuned) drops from 17.6 to 8.8 at $L_3$ and from 11.8 to 2.6 at $L_4$, while Uni-NaVid (fine-tuned) drops from 5.1 to 3.8 at $L_3$. This repeated degradation indicates that current architectures lack sufficiently stable spatial memory reasoning for compositional return navigation.

\vspace{-0.2cm}
\subsection{Behavioral-to-Architectural Analysis}
\begin{figure}[tb]

    \centering
    \includegraphics[width=1\linewidth]{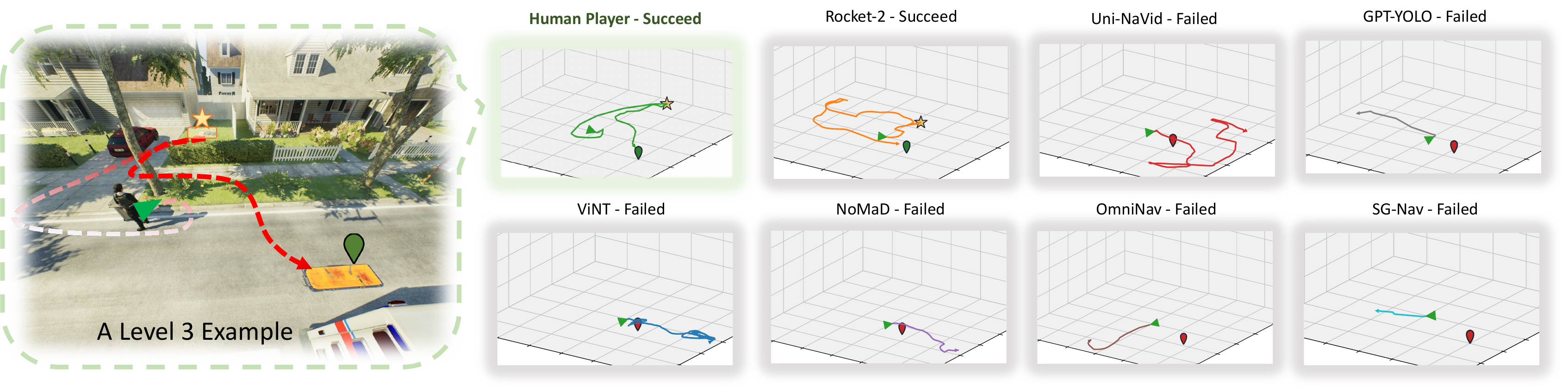}
    \vspace{-0.5cm}
    \caption{Trajectory comparison on a representative $L_3$
  episode across four architectural paradigms. \textbf{Modular} (LLM-YOLO Planner): lacking obstacle avoidance, the agent becomes trapped by obstacles and idles. \textbf{Map-based} (SG-Nav, OmniNav): inference latency yields minimal spatial coverage. \textbf{VLN} (ViNT, NoMaD): degenerate looping exhausts the time budget. Uni-NaVid and ROCKET-2: systematic exploration with active search coverage, yet imperfect terminal convergence.}
    \label{fig:Traj_compare}
\end{figure}

These stage-level results indicate \emph{where} methods fail. To understand \emph{why}, we turn to trajectory-level behavior. Figure~\ref{fig:Traj_compare} compares trajectories across paradigms on the same representative $L_3$ episode, exposing qualitatively distinct failure patterns that trace back to architectural design choices. We observe that:

~\textbf{(1) Modular detection} (LLM-YOLO Planner) collapses at $L_2$: a single perception failure in the cascaded pipeline propagates irreversibly with no recovery mechanism, causing the agent to fail within the critical survival time window. 

~\textbf{(2) Map-based models} (SG-Nav, OmniNav) are bottlenecked by inference latency---at 0.2--0.4 FPS, only dozens of steps fit within the 300-second budget, far too few for meaningful spatial coverage. 

~\textbf{(3) VLN models} (ViNT, NoMaD) exhibit degenerate looping that exhausts the time budget without exploring new areas. This reflects a fundamental architectural constraint: topological graph-based route retrieval yields no matches in unfamiliar environments, leaving the policy to follow raw predictions without exploration logic. Fine-tuning breaks the zero-completion barrier (TCR: 0\% to 8\% at $L_1$), confirming that the bottleneck is the route-following paradigm rather than model capacity; moreover, NoMaD's diffusion U-Net consistently outperforms ViNT's deterministic FC layer (e.g., fine-tuned $L_4$ S1: 6.4 vs.\ 4.5), indicating that modeling multimodal action distributions enables more systematic exploration.

~\textbf{(4) VLA model and goal-conditioned visuomotor policy} (Uni-NaVid, ROCKET-2) display exploration-competent trajectories with active search patterns, confirming that the end-to-end temporal architecture intrinsically supports open-world exploration. Their residual failures stem from imperfect terminal convergence and weak spatial memory, causing time-outs during inefficient backtracking.

Together, these patterns indicate that the exploration deficit is not a data or capacity issue but an architectural one: only paradigms with intrinsic exploration logic (VLA, goal-conditioned visuomotor policies) achieve systematic coverage, while route-retrieval and cascaded architectures fundamentally lack this capability.

\vspace{-0.2cm}
\section{Conclusion}
\vspace{-0.2cm}
We introduce RescueBench, which reveals a clear gap between current embodied agents and realistic multi-stage search-and-rescue tasks. By unifying photo-realistic simulation, progressive difficulty, sequential task composition, and diverse embodied baselines, RescueBench provides a diagnostic benchmark for studying exploration, memory, and goal-conditioned navigation in embodied SAR. Across five progressive difficulty levels, all evaluated methods fail to complete the full four-stage pipeline at L5. Stage-level analysis shows that failures primarily arise from exploration under uncertainty and are further amplified by weak spatial memory during return navigation. These results suggest that the observed limitations are architectural rather than merely data-driven within the evaluated settings. 


\bibliographystyle{unsrt}

\bibliography{main}
\appendix
\newpage

\section*{Appendix}
This appendix is organized into three parts. Section~\ref{app:results_ext} extends the experimental evidence from the main text with complete quantitative results, fine-tuning details, and supplementary trajectory visualizations. Section~\ref{app:benchmark_details} documents benchmark construction, including environment diversity, difficulty design, the automatic data collection pipeline, and the simulator interface and evaluation protocol. Section~\ref{app:metric_details} provides the definition and diagnostic interpretation of the Human Similarity metric.

\section{Related Work}

\textbf{Vision-based embodied navigation.} Vision-language navigation (VLN) \cite{anderson2018vision,fried2018speaker,krantz2020beyond} grounds navigation in natural language instructions, yet the instruction-following paradigm fundamentally differs from the open-ended exploration that SAR demands: VLN agents are told where to go, whereas SAR agents must decide where to go based on ambiguous multimodal clues. Navigation foundation models such as ViNT \cite{shah2023vint} and NoMaD \cite{sridhar2024nomad} learn generalizable visual navigation representations through large-scale pretraining, but both rely on topological graph retrieval for waypoint selection, an architectural constraint that fundamentally limits open-world exploration. Goal-conditioned visuomotor policies such as ROCKET-2 \cite{cai2025rocket} and video-based VLA models such as Uni-NaVid \cite{zhang2024uni} demonstrate zero-shot transfer across diverse 3D environments, yet their residual gaps in precise terminal convergence and spatial memory point to specific capabilities that future architectures must strengthen.

\textbf{Map-based navigation.} Methods such as SG-Nav \cite{yin2024sg} and OmniNav leverage explicit spatial representations for navigation planning. SG-Nav constructs online 3D scene graphs for LLM-based zero-shot reasoning, while OmniNav employs vision-language grounding over pre-built environment maps. Their structured spatial reasoning is theoretically well-suited to SAR, yet per-step inference latency (0.2--0.4 FPS) limits effective exploration coverage under step budgets, revealing a deployment-efficiency gap distinct from the exploration-strategy gap of VLN models.

\textbf{Embodied AI benchmarks.} Navigation benchmarks such as R2R \cite{anderson2018vision} and VLN-CE \cite{krantz2020beyond} standardized instruction-following navigation but remain confined to indoor environments. Outdoor extensions including CityWalker \cite{liu2025citywalker} and UrbanNav \cite{mei2026urbannav} expand spatial scale yet assume known destinations, eliminating the autonomous exploration prerequisite for SAR. Manipulation benchmarks \cite{liu2023libero,mees2022calvin} evaluate tabletop interaction but decouple manipulation from large-scale navigation. Integrated indoor benchmarks such as ALFRED \cite{shridhar2020alfred} and Habitat~3.0 \cite{puig2023habitat} combine navigation and manipulation, yet none provide progressive difficulty levels or support heterogeneous multi-agent evaluation. Recent multi-agent benchmarks including PARTNR \cite{changpartnr} and COHERENT \cite{liu2025coherent} evaluate collaborative task decomposition, but focus on LLM-based discrete action planning in structured indoor settings rather than continuous sensorimotor control. Table~\ref{tab:benchmark_comparison} summarizes the capability landscape. Rather than merely combining exploration, interaction, and memory into one scenario, RescueBench's primary contribution is enabling diagnosis of how failures propagate across sequentially dependent embodied sub-tasks. Existing benchmarks evaluate these capabilities in isolation or in parallel; RescueBench exposes cascading failure modes that only emerge under sequential composition. The benchmark additionally provides five progressive difficulty levels and an extensible framework for heterogeneous multi-agent evaluation, with preliminary validation in this submission.


\section{Extended Results and Additional Analysis}
\label{app:results_ext}
This section consolidates the quantitative and qualitative evidence that supports the main experimental claims, including complete result tables, fine-tuning details.

\subsection{Overall Quantitative Results}
This subsection reports the complete overall results underlying the summary trends discussed in the main text, including zero-shot, fine-tuned, and human reference performance across all five difficulty levels, as shown in Table ~\ref{tab:overall_zeroshot}.

\begin{table}[tbp!]
\centering
\caption{Overall performance across five difficulty levels, covering all baseline methods and their fine-tuned version, followed by ``(fine-tuned)''. Each cell reports \textbf{TCR (\%)} / \textbf{TS (0--100)} / \textbf{Avg Time Cost (s)} / \textbf{Avg Collision Counts}.} 
\label{tab:overall_zeroshot}
\resizebox{\textwidth}{!}{%
\begin{tabular}{lc|c|c|c|c|c}
\toprule
 & &\multicolumn{5}{c}{\textbf{TCR (\%) / TS (0--100) / Avg Time Cost (s)/Avg Collision (counts)}} \\
\textbf{Method} & \textbf{FPS} & \textbf{L1} & \textbf{L2} & \textbf{L3} & \textbf{L4} & \textbf{L5} \\
\midrule
LLM-YOLO Planner & 3.3
  & 48.0/72.2/91.5/23.8
  & 0.0/12.1/175.9/67.6
  & 0.0/5.7/237.2/96.2
  & 0.0/6.0/300.4/147.9
  & 0.0/6.2/300.5/152.6 \\
Uni-NaVid         & 6.2
  & 24.0/67.4/103.4/27.8
  & 11.8/21.9/165.5/45.6
  & 5.1/14.2/234.8/68.1
  & 2.7/12.9/286.8/120.2
  & 0.0/12.2/300.3/86.1 \\
ViNT            & 6.8
  & 0.0/14.2/180.2/172.2
  & 0.0/8.2/180.2/110.6
  & 0.0/4.7/240.2/199.2
  & 0.0/4.3/300.2/272.0
  & 0.0/1.8/300.2/274.6 \\ 
NoMaD           & 5.6
  & 0.0/10.0/180.2/162.0
  & 0.0/6.4/180.3/111.0
  & 0.0/3.9/240.3/155.1
  & 0.0/7.0/300.3/226.3
  & 0.0/5.8/300.5/229.5 \\
SG-Nav           &0.2  
  & 0.0/62.3/198.5/0.0 
  & 0.0/14.1/199.4/0.7
  & 0.0/8.4/240.3/0.7  
  & 0.0/10.0/330.5/1.4
  & 0.0/2.0/323.9/0.8 \\
OmniNav(Visual)         & 0.4
  & 0.0/15.6/176.6/4.6
  & 0.0/11.0/184.9/3.4
  & 0.0/4.5/240.4/3.8
  & 0.0/6.8/305.0/5.9
  & 0.0/3.3/304.8/8.4 \\
ROCKET-2                &4.6
 &16.0/59.0/125.3/20.6
 &14.7/57.2/130.2/25.4
 &10.0/42.6/193.4/54.2
 &2.5/32.2/254.7/84.2
 &0.0/8.2/296.7/106.9\\ \hline
Uni-NaVid(fine-tuned) &6.0
&52.0/80.9/66.5/21.6
&11.8/26.1/166.4/62.8
&11.9/18.2/221.3/61.0 
&10.1/27.3/275.5/86.3
&0.0/7.0/300.3/132.3\\
ViNT (fine-tuned)             &5.5
  &8.0/18.2/172.0/66.9	
  &3.4/10.5/170.8/94.2	
  &1.6/8.4/236.9/148.5	
  &0.0/6.6/300.3/193.9	
  &0.0/2.8/300.3/219.8 \\
NoMaD (fine-tuned)          &5.1
&8.0/25.3/175.2/9.2
&1.7/11.3/179.4/21.1
&0.0/3.2/240.3/29.5
&0.0/11.5/300.3/49.3 
&0.0/1.0/300.3/49.0 \\ 
ROCKET-2 (fine-tuned)         & 4.5
  & 32.0/66.0/102.8/11.0
  & 11.8/66.8/125.2/20.6
  & 15.0/52.0/192.6/50.1
  & 5.2/29.0/252.5/99.9
  & 0.0/4.5/300.3/103.7 \\ 
\midrule
Human Player     & 5.1
  &100.0/100.0/19.2/0.0  & 100.0/100.0/47.1/0.0 &100.0/100.0/56.3/2.1 & 100.0/100.0/76.4/10.3 & 93.5/95.2/107.8/24.2 \\
Oracle &9.9
&100.0/100.0/15.2/0.0
&100.0/100.0/34.5/16.9
&100.0/100.0/45.2/17.2
&100.0/100.0/55.3/60.3
&100.0/100.0/81.7/83.2\\
\bottomrule
\end{tabular}%
}
\end{table}
\subsection{Stage-Level Diagnosis}
This subsection provides the full per-stage breakdown for all methods, making explicit how failures propagate across exploration, rescue, return, and handoff as difficulty increases, as shown in Table ~\ref{tab:stage_all_1} and Table ~\ref{tab:stage_all_2}.

\begin{table}[tbp!]
\centering
\caption{
Per-stage capability diagnosis (\textbf{Part I: L1--L3}). Each row reports \textbf{Stage-TS (/25)} for S1 = Explore, S2 = Rescue, S3 = Return, S4 = Handoff. \textbf{Avg Steps}: average number of executed steps. \textbf{Human Similarity} (HS): behavioral similarity between agents and human players.
}
\label{tab:stage_all_1}
\resizebox{\textwidth}{!}{%
\begin{tabular}{c|l|c|c|c}
\toprule
\textbf{Level} & \textbf{Method} & \textbf{StageScore (S1/S2/S3/S4)} & \textbf{AVG Steps} & \textbf{Human Similarity (\%)} \\
\midrule

\multirow{9}{*}{L1}
 & LLM-YOLO Planner  & 23.0/20.3/15.0/13.9 & 343.6  & 80.5 \\
 & Uni-NaVid         & 19.0/16.8/16.0/15.6 & 796.0  & 82.7 \\
 & ViNT              & 9.0/4.2/1.0/0.0     & 1520.6 & 51.6 \\
 & NoMaD             & 5.0/5.0/0.0/0.0     & 1287.6 & 33.4 \\
 & SG-Nav            & 24.0/20.4/12.0/6.0  & 40.6   & 92.4 \\
 & OmniNav           & 7.0/4.7/2.0/1.9     & 155.5  & 42.4 \\
  & ROCKET-2         & 18.0/17.2/13.0/10.8&763.0 &65.2 \\ 
& Uni-NaVid (fine-tuned) &21.0/21.0/20.0/18.9&517.0&84.9 \\
& ViNT (fine-tuned) &9.0/5.2/2.0/2.0&1063.0&43.4 \\
& NoMaD (fine-tuned) &12.0/8.0/3.0/2.2&911.0& 70.8\\
 & ROCKET-2   (fine-tuned)        & 18.0/18.0/16.0/14.0 & 575.8  & 69.2 \\
\cmidrule{2-5}
 & Human Player      & 25.0/25.0/25.0/25.0 & 111.1  & 100.0 \\
  & Oracle &25.0/25.0/25.0/25.0&197.2&97.8 \\
\midrule

\multirow{9}{*}{L2}
 & LLM-YOLO Planner  & 6.6/5.5/0.0/0.0     & 574.3  & 43.4 \\
 & Uni-NaVid         & 8.1/7.9/2.9/2.9     & 956.9  & 53.4 \\
 & ViNT              & 3.6/2.5/1.5/0.5     & 111.1  & 53.4 \\
 & NoMaD             & 3.6/2.7/0.0/0.0     & 956.1  & 47.9 \\
 & SG-Nav            & 8.1/5.3/0.7/0.1     & 37.2   & 73.6 \\
 & OmniNav           & 5.1/3.3/1.5/1.5     & 74.8   & 58.4 \\
& ROCKET-2         & 19.8/19.0/9.6/8.8&527.5 & 60.9\\ 
& Uni-NaVid (fine-tuned) &10.3/8.9/3.7/3.1&958.5&55.3 \\
& ViNT (fine-tuned) & 3.8/3.3/1.7/1.7 &882.0 &56.0 \\
& NoMaD (fine-tuned) &5.5/3.5/1.7/0.6&898.6& 58.1\\
 & ROCKET-2 (fine-tuned)          & 21.3/20.5/13.9/11.0 & 496.9  & 64.7 \\
\cmidrule{2-5}
 & Human Player      & 25.0/25.0/25.0/25.0 & 201.4  & 100.0 \\
  & Oracle &25.0/25.0/25.0/25.0&292.1&93.1 \\
\midrule

\multirow{9}{*}{L3}
 & LLM-YOLO Planner  & 3.8/1.9/0.0/0.0     & 748.0  & 43.9 \\
 & Uni-NaVid         & 5.4/4.4/2.5/1.9     & 1366.2 & 35.5 \\
 & ViNT              & 3.4/1.3/0.0/0.0     & 1584.8 & 50.5 \\
 & NoMaD             & 2.5/1.3/0.0/0.0     & 1321.7 & 40.6 \\
 & SG-Nav            & 5.1/2.9/0.4/0.0     & 43.4   & 72.3 \\
 & OmniNav           & 4.1/2.4/0.0/0.0     & 80.0   & 60.7 \\
 & ROCKET-2          & 15.0/13.3/7.5/6.8& 807.9&42.2 \\ 
 & Uni-NaVid (fine-tuned) &5.5/5.1/3.8/3.8&1215.1&44.5\\
& ViNT (fine-tuned)       &4.7/2.9/0.4/0.4&1322.0&57.5    \\
& NoMaD (fine-tuned)       &1.7/1.6/0.0/0.0&1185.1&55.0\\
 & ROCKET-2 (fine-tuned)  & 18.4/17.6/8.8/7.4   & 804.4  & 42.5 \\
\cmidrule{2-5}
\cmidrule{2-5}
 & Human Player      & 25.0/25.0/25.0/25.0 & 260.4  & 100.0 \\
 & Oracle             &25.0/25.0/25.0/25.0&410.8&92.0 \\
\midrule
\end{tabular}%
}
\end{table}

\begin{table}[tbp!]
\centering
\caption{Per-stage capability diagnosis (\textbf{Part II: L4--L5 and average}). Each row reports \textbf{Stage-TS (/25)} for S1 = Explore, S2 = Rescue, S3 = Return, S4 = Handoff. \textbf{Avg Steps}: average number of executed steps under the episode time budget. \textbf{Human Similarity} (HS) reflects the behavioral similarity between human players and embodied agents.}
\label{tab:stage_all_2}
\resizebox{\textwidth}{!}{%
\begin{tabular}{c|l|c|c|c}
\toprule
\textbf{Level} & \textbf{Method} & \textbf{StageScore (S1/S2/S3/S4)} & \textbf{AVG Steps} & \textbf{Human Similarity (\%)} \\
\midrule
\midrule

\multirow{9}{*}{L4}
 & LLM-YOLO Planner  & 3.8/1.7/0.0/0.0     & 973.6  & 50.1 \\
 & Uni-NaVid         & 5.8/4.6/1.3/1.2     & 1695.0 & 51.6 \\
 & ViNT              & 3.2/1.1/0.0/0.0     & 1989.4 & 65.4 \\
 & NoMaD             & 4.5/2.5/0.0/0.0     & 1580.3 & 51.0 \\
 & SG-Nav            & 6.4/3.6/0.0/0.0     & 47.8   & 61.7 \\
 & OmniNav           & 4.5/2.4/0.0/0.0     & 115.2  & 53.6 \\
 & ROCKET-2          &11.5/11.5/5.1/4.0 &1179.3&59.7 \\
 & Uni-NaVid (fine-tuned)  &9.0/8.5/5.1/4.6 &1504.7&48.2 \\
& ViNT (fine-tuned)        & 4.5/2.1/0.0/0.0  & 1468.8 &53.3 \\
& NoMaD (fine-tuned)        &6.4/5.1/0.0/0.0&1558.3&72.7 \\
 & ROCKET-2 (fine-tuned)         & 12.2/11.8/2.6/2.4& 1194.0 & 59.4 \\
\cmidrule{2-5}
\cmidrule{2-5}
 & Human Player      & 25.0/25.0/25.0/25.0 & 381.6  & 100.0 \\
  & Oracle         &25.0/25.0/25.0/25.0&565.7&90.1 \\
\midrule

\multirow{9}{*}{L5}
 & LLM-YOLO Planner  & 4.3/1.1/0.0/0.0     & 985.7  & 56.0 \\
 & Uni-NaVid         & 7.3/4.9/0.0/0.0     & 1851.1 & 53.3 \\
 & ViNT              & 1.6/0.2/0.0/0.0     & 2072.2 & 58.7 \\
 & NoMaD             & 4.8/1.0/0.0/0.0     & 1539.5 & 51.8 \\
 & SG-Nav            & 1.6/0.3/0.0/0.0     & 48.0   & 47.4 \\
 & OmniNav           & 2.4/0.8/0.0/0.0     & 125.1  & 48.7 \\
 & ROCKET-2          &5.6/2.6/0.0/0.0      &1538.3&48.7 \\
 & Uni-NaVid (fine-tuned) &4.0/3.0/0.0/0.0&1828.2&47.2 \\
& ViNT (fine-tuned)       & 2.4/0.4/0.0/0.0  &1794.2  &56.3 \\
& NoMaD (fine-tuned)       &0.8/0.2/0.0/0.0&1606.2&69.8 \\
 & ROCKET-2 (fine-tuned)  & 2.4/2.1/0.0/0.0     & 1523.6  & 48.4 \\
\cmidrule{2-5}
\cmidrule{2-5}
 & Human Player      & 24.2/24.2/23.4/23.4 & 555.0  & 100.0 \\
  & Oracle &25.0/25.0/25.0/25.0&869.9&86.4 \\
 \midrule

\multirow{9}{*}{Avg. L1-L5}
 & LLM-YOLO Planner  & 7.0/4.9/2.0/1.8     & 748.8  & 52.0 \\
 & Uni-NaVid         & 8.1/6.8/3.7/3.5     & 1364.5 & 51.3 \\
 & ViNT              & 3.9/1.7/0.4/0.1     & 1654.9 & 55.6 \\
 & NoMaD             & 5.0/5.0/0.0/0.0     & 1340.7 & 44.9 \\
 & SG-Nav            & 7.8/4.4/1.9/0.8     & 43.5   & 68.9 \\
 & OmniNav           & 4.1/2.4/0.5/0.5     & 103.7  & 54.3 \\
 & ROCKET-2          &14.0/12.7/6.9/6.0    &948.0& 53.3\\
 & Uni-NaVid (fine-tuned) &8.9/8.3/5.6/5.2&1237.0&53.0 \\
& ViNT (fine-tuned)  &4.9/3.0/1.1/0.8&1265.2 & 54.3\\
& NoMaD (fine-tuned) &4.7/3.3/0.8/0.4&1203.2& 63.8\\
 & ROCKET-2 (fine-tuned)         & 14.9/14.4/7.9/6.7   & 917.2  & 54.5 \\
\cmidrule{2-5}
\cmidrule{2-5}
 & Human Player      & 24.9/24.9/24.7/24.7 & 291.6  & 100.0 \\
  & Oracle &25.0/25.0/25.0/25.0&468.25&91.7 \\
\bottomrule

\end{tabular}%
} 
\end{table}

\subsection{Fine-Tuning with Automatically Collected Data}
\label{app:finetune}
Among the evaluated baselines, we fine-tune the four methods that support RescueBench-specific adaptation: Uni-NaVid, ViNT, NoMaD, and ROCKET-2. All training data are obtained through the automatic data collection pipeline proposed in this paper, allowing us to directly test whether the collected demonstrations provide effective supervision. Keeping the remaining baselines in zero-shot mode preserves a clean comparison and reveals which architectural families can translate RescueBench-specific data into measurable gains.

All fine-tuning data are obtained from the same automatic pipeline described in Appendix~\ref{app:pipeline}, with approximately 400K automatically collected expert steps in total. We adopt architecture-specific adaptation strategies. For Uni-NaVid, we continue training for 2 epochs with an effective batch size of 64 (per-device batch size 16 with gradient accumulation 4), AdamW-style optimization ($\beta_1=0.9,\ \beta_2=0.999$), a learning rate of $1\times10^{-5}$, cosine scheduling, and a warmup ratio of 0.03. For ViNT and NoMaD, we fine-tune the pretrained checkpoints with AdamW ($\beta_1=0.9,\ \beta_2=0.999$), cosine decay, and an initial learning rate of $1\times10^{-4}$. ViNT is trained for 5 epochs with batch size 256, while NoMaD is trained for 30 epochs with batch size 64, following the official fine-tuning recipe where applicable. For ROCKET-2, we follow the official training configuration during fine-tuning. We use a predefined 9:1 train/validation split in the processed Rescue training set, together with model-specific input resolutions (ViNT: $85\times64$, NoMaD: $96\times96$, and Uni-NaVid processed with the CLIP-224 image pipeline). All fine-tuning experiments are conducted on a single NVIDIA RTX 5090D.

The results in Table~\ref{tab:overall_zeroshot}, Table ~\ref{tab:stage_all_1} and Table ~\ref{tab:stage_all_2} show a consistent pattern. At $L_1$, Uni-NaVid and ROCKET-2 both double their TCR, while ViNT and NoMaD break through from 0\% to 8\%. This confirms that the automatically collected data are useful as supervision, but also that the benefit is architecture-dependent: data alone can improve exploration quality, yet it does not remove the underlying bottlenecks of topological graph retrieval or weak spatial memory.

\section{Benchmark Construction Details}
This section documents how RescueBench is constructed, from baseline method details, environment selection and difficulty design, to automatic episode generation and evaluation-time simulator settings.
\label{app:benchmark_details}

\subsection{Baseline Method Details}
\label{app:baseline_details}
This subsection provides detailed descriptions of the seven evaluated baselines and two non-learning references, organized by architectural type.

\paragraph{Classical Two-Stage.} \textbf{LLM-YOLO Planner} combines a YOLO-based victim detector with a high-level planner (GPT-4o). The modular two-stage architecture, perception followed by language-guided action selection, provides an interpretable baseline while also revealing the brittleness of cascaded pipelines when perception degrades under visual complexity.

\paragraph{Video VLA.} \textbf{Uni-NaVid} \cite{zhang2024uni} unifies multiple embodied navigation tasks through end-to-end video-conditioned action prediction, leveraging temporal reasoning over video sequences for spatial-temporal understanding.

\paragraph{VLN Foundation Models.} \textbf{ViNT} \cite{shah2023vint} and \textbf{NoMaD} \cite{sridhar2024nomad} represent navigation foundation models that learn generalizable sensorimotor policies from large-scale demonstration data. ViNT learns waypoint-following behaviors, while NoMaD generates diverse goal-directed trajectories via a goal-masking diffusion policy. Both operate in an end-to-end manner, relying on topological graph retrieval to select waypoints, an architectural constraint that fundamentally limits open-world exploration.

\paragraph{Map-Based Methods.} \textbf{SG-Nav} \cite{yin2024sg} and \textbf{OmniNav} \cite{xue2026omninav} integrate explicit spatial representations for navigation planning. SG-Nav constructs online 3D scene graphs for LLM-based zero-shot navigation, while OmniNav employs a vision-language grounding pipeline over pre-built environment maps. Both reason over structured spatial representations rather than learning end-to-end sensorimotor policies, but their per-step inference latency (0.2--0.4 FPS) limits effective exploration coverage under step budgets.

\paragraph{Goal-Conditioned Visuomotor Policy.} \textbf{ROCKET-2} \cite{cai2025rocket} steers navigation via cross-view goal alignment with native interaction capabilities (pick, drop, open door), the only evaluated baseline whose architecture natively supports all four SAR stages without external interaction modules.

\paragraph{Reference Baselines.} Two human players unfamiliar with the environments are randomly assigned task instances across all five difficulty levels, using the same keyboard/mouse interface. The \textbf{Oracle Navigator} is a non-learning upper bound provided with ground-truth environment information (target location, ambulance location, obstacle-free shortest path) at each episode start, executing the full four-stage pipeline via the simulator's built-in navigation with a hand-crafted state machine. By removing exploration uncertainty and perception errors entirely, the Oracle quantifies the execution-time lower bound, revealing how much difficulty stems from search-and-detection versus pure locomotion and interaction overhead.

\subsection{Testbed Diversity and Distribution}
This subsection summarizes the scale and diversity of the RescueBench test set, which contains 213 episodes spanning seven environments and five difficulty levels. Although the episode count is modest, the spatial scale of each episode is substantial: the Oracle agent accumulates approximately \textbf{7.8\,km} of traversed paths across all 213 episodes, with individual episode trajectories ranging from roughly 15\,m (short $L_1$ episodes) to over 150\,m (challenging $L_5$ episodes). Table~\ref{tab:oracle_path} further breaks down the cumulative Oracle path length by difficulty level, confirming that spatial demands are substantial across all levels rather than concentrated at a single difficulty. This per-episode spatial extent ensures that each method is evaluated on a diverse and geographically extensive set of navigation challenges, and that the cumulative path length far exceeds typical indoor navigation benchmarks. RescueBench spans diverse environments that vary along indoor and outdoor layout, single-floor and multi-floor structure, and architectural type. The figures below illustrate both the scene diversity and the resulting statistical profile of the test set.

\begin{table}[tbp!]
\centering
\caption{Cumulative Oracle path length by difficulty level. The Oracle navigates obstacle-free shortest paths for the full four-stage pipeline (explore $\to$ rescue $\to$ return $\to$ handoff), providing a lower bound on the spatial distance each agent must traverse.}
\label{tab:oracle_path}
\begin{tabular}{l c}
\toprule
\textbf{Difficulty Level} & \textbf{Cumulative Path Length (m)} \\
\midrule
$L_1$ & 354.8 \\
$L_2$ & 1,578.0 \\
$L_3$ & 2,131.0 \\
$L_4$ & 1,752.6 \\
$L_5$ & 1,995.6 \\
\midrule
\textbf{Total} & \textbf{7,812.0} \\
\bottomrule
\end{tabular}
\end{table}

\begin{figure}[tbp!]
    \centering
    \caption{ Overview of the seven virtual environments in RescueBench. \textbf{FlexibleRoom} (top-left in Fig ~\ref{fig:progress_diff}) is a clean indoor scene serving as the basic capability baseline. The remaining six environments are photo-realistic, covering daily residential, commercial street, and outdoor natural scenes. This diversity ensures that evaluation results generalize beyond a single visual domain.
    }
    \includegraphics[width=1\linewidth]{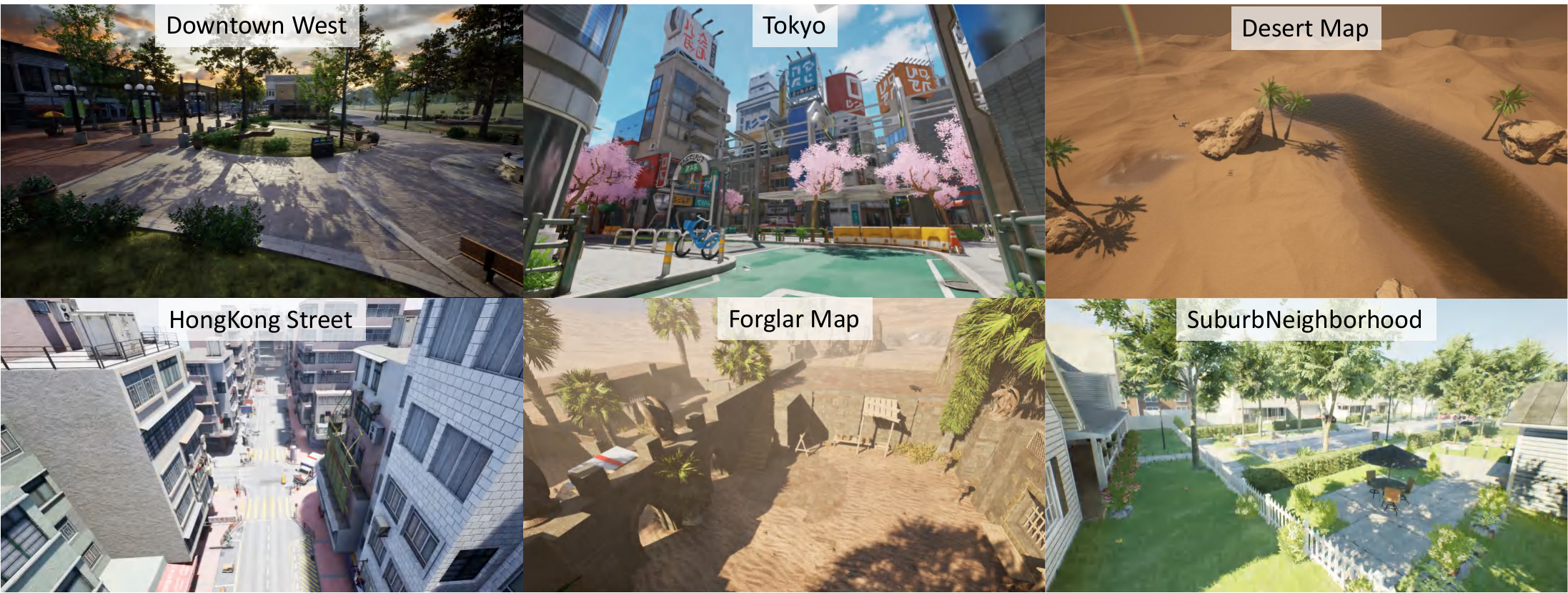}
    \label{fig:env_gallery}
\end{figure}

\begin{figure}[tbp!]
    \centering
    \begin{subfigure}{\linewidth}
        \centering
        \includegraphics[width=\linewidth]{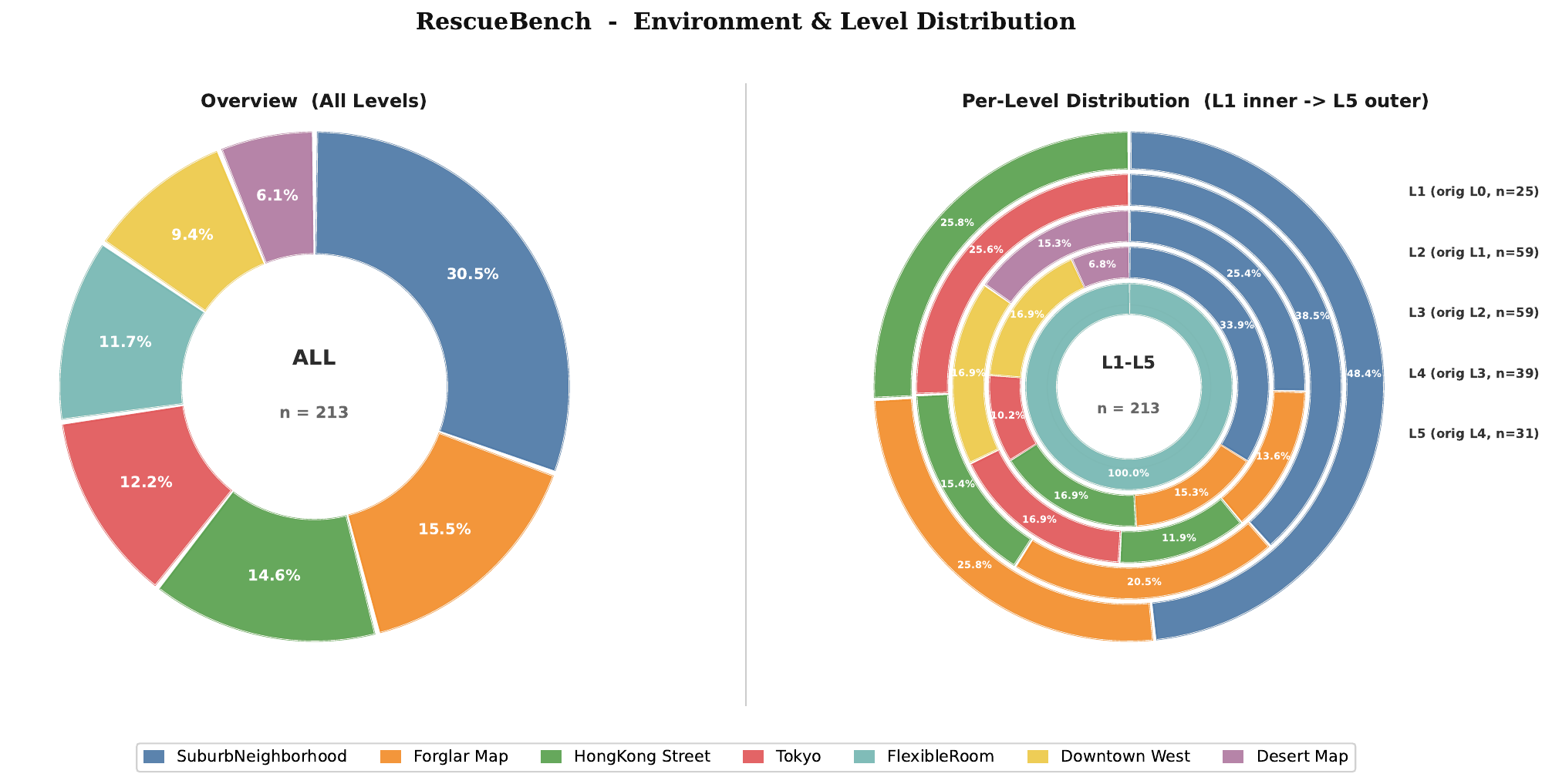}
        \label{fig:testset_stats}
    \end{subfigure}
    \vspace{4pt} 
    \begin{subfigure}{\linewidth}
        \centering
        \includegraphics[width=\linewidth]{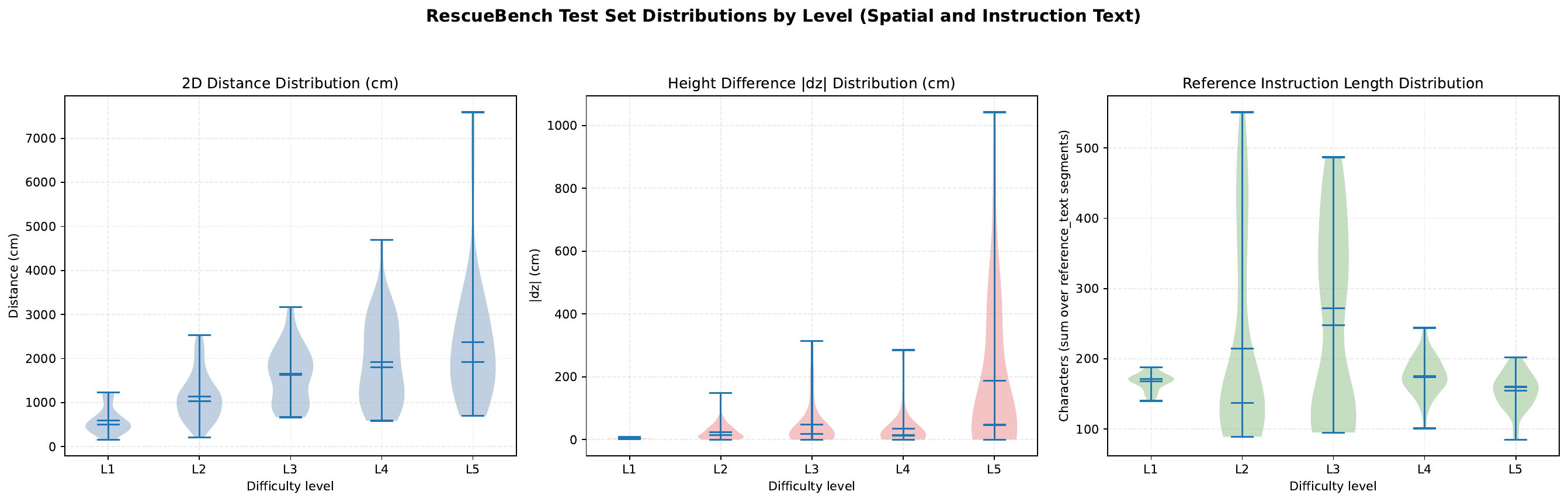}
        \label{fig:testset_distance}
    \end{subfigure}
    \caption{Statistical summary of the RescueBench test set ($n=213$ episodes). The top panel shows the distribution of environments and difficulty levels. The bottom panel shows per-level violin plots for three quantities induced by the sampling rules: target-to-ambulance 2D Euclidean distance, height difference $|\Delta z|$, and reference instruction length (characters).}
    \label{fig:testset_overview}
\end{figure}

\section{Difficulty Level Specification}
\label{app:difficulty}
The five difficulty levels $\mathcal{L} = \{L_1, \ldots, L_5\}$ are defined through a combination of generative constraints during episode sampling and validation filters applied after generation. Table~\ref{tab:difficulty_spec} summarizes the per-level specification.

\begin{table}[tbp!]
\centering
\caption{Difficulty level specification. Each level is defined by a distance range $[d_{\min}^{(l)}, d_{\max}^{(l)}]$ for target placement and two validation criteria. $\checkmark$ = required, $-$ = must not occur, $\sim$ = may occur but not required.}
\label{tab:difficulty_spec}
\begin{tabular}{l|c|c|c|c}
\toprule
\textbf{Level} & \textbf{$d_{\min}$ (m)} & \textbf{$d_{\max}$ (m)} & \textbf{Height $\Delta z$} & \textbf{Interaction} \\
\midrule
$L_1$ & 5 & 15 & $-$ & $-$ \\
$L_2$ & 5 & 30 & $-$ & $-$ \\
$L_3$ & 10 & 40 & $\sim$ & $-$ \\
$L_4$ & 10 & 60 & $\sim$ & $\checkmark$\\
$L_5$ & 10 & 80 & \textbf{$\checkmark$} & $\checkmark$ \\
\bottomrule
\end{tabular}
\end{table}

\textbf{Per-level progression.} $L_1$ and $L_2$ both place the target at close range and differ primarily in environmental appearance: $L_1$ uses a clean open platform, whereas $L_2$ introduces cluttered outdoor backgrounds that challenge perception without requiring autonomous search. $L_3$ marks the key discontinuity, moving the target to a longer initial distance range ($d_{\min}=10$\,m) and requiring short-range exploration within a single area. $L_4$ further extends the range and adds navigable interaction elements such as doors and gates that force the agent to cross region boundaries. $L_5$ adds 3D structural complexity on top of $L_4$, with a minimum elevation-change requirement that creates multi-floor layouts.

\textbf{Clue ambiguity progression.} Text clues are generated by Qwen-3.5 Plus: ego-centric video clips from the expert trajectory are fed to the model together with a level-specific prompt, producing natural-language reference instructions. The Clue ambiguity escalates in parallel with environmental complexity. $L_1$--$L_2$ clues supplement the close-range visual signal without carrying the search burden. $L_3$ clues provide rich visual detail (landmarks, clothing, scene layout) that the agent must match to observations during search. $L_4$ clues shift to route-level directional language requiring cross-region wayfinding, and $L_5$ clues become deliberately sparse and hierarchical, forcing agents to decompose a minimal spatial goal across multiple floors without step-by-step guidance. This progression from visual matching through directional reasoning to hierarchical decomposition mirrors real-world SAR information regimes. Figure~\ref{fig:testset_overview} (bottom panel, right) visualizes the per-level distribution of reference instruction lengths.

\textbf{Distance constraints.} During episode configuration (Section~3.4), the target position $\mathbf{p}_{\text{tar}}$ is sampled such that the Euclidean distance to the ambulance satisfies $d(\mathbf{p}_{\text{amb}}, \mathbf{p}_{\text{tar}}) \in [d_{\min}^{(l)}, d_{\max}^{(l)}]$ for the target level $l$. The ranges are intentionally overlapping across adjacent levels: $L_1$ and $L_2$ share the same lower bound ($d_{\min}=5$\,m) but $L_2$ extends to farther distances; $L_3$--$L_5$ shift the lower bound to $d_{\min}=10$\,m to place the target beyond immediate visual range, with progressively larger $d_{\max}$. The resulting target-ambulance distance distributions are shown in Figure~\ref{fig:testset_overview} (bottom panel, left). The agent's initial position is sampled uniformly within 7.5\,m of $\mathbf{p}_{\text{amb}}$.

\textbf{Validation filters.} After sampling, each episode passes through a two-criterion filter (Section~3.4) that guarantees correspondence between the sampled configuration and the labeled difficulty level:

\begin{enumerate}
    \item \textbf{3D elevation change ($\Delta z$).} For $L_5$, the expert trajectory must have a minimum elevation change $\Delta z_{\min}$ (approximately one full floor height, 3--4\,m), ensuring that the agent must traverse staircases, ramps, or multi-level platforms. For $L_1$--$L_2$, episodes with significant $\Delta z$ are rejected, keeping these levels on flat terrain. For $L_3$--$L_4$, moderate elevation changes may occur but are not enforced. The resulting height distributions are shown in Figure~\ref{fig:testset_overview} (bottom panel, center).
    \item \textbf{Environment interaction.} For $L_4$--$L_5$, the expert trajectory must contain at least one navigable interaction element (if the interaction exists in the environment): a door, gate, or narrow passageway that requires the agent to actively engage with the environment structure. This ensures that these levels test interaction-conditional navigation rather than open-space traversal alone. $L_1$--$L_3$ are deliberately free of interaction constraints, isolating spatial reasoning from physical interaction demands.
\end{enumerate}

All episodes failing any applicable criterion are rejected and re-sampled until a valid configuration is found, guaranteeing that every collected episode faithfully represents its labeled difficulty. This generative-constraint-plus-validation design ensures both scalability (unlimited validated episodes per environment) and faithful difficulty correspondence without manual curation.

\subsection{Automatic Data Collection Pipeline}
\label{app:pipeline}

This section provides the full implementation details of the four-stage automatic data collection pipeline referenced in Section~3.4.

\paragraph{(1) Navigable region sampling.} The pipeline exports $\mathcal{M}_{\text{nav}}$ from the target environment via a custom UE5 plugin and performs uniform sampling over all traversable regions to construct a candidate point pool $\mathcal{P}$. To ensure that ambulance placement is physically feasible (given the large footprint of vehicles and stretchers), $\mathcal{P}$ is partitioned into contiguous regions, and the largest region without narrow branches is designated as the home base candidate zone $\mathcal{Z}_{\text{amb}}$.

\paragraph{(2) Constrained episode configuration.} For each episode, the ambulance position $\mathbf{p}_{\text{amb}}$ is sampled uniformly from $\mathcal{Z}_{\text{amb}}$, and the injured person position $\mathbf{p}_{\text{tar}}$ is sampled from $\mathcal{P}$ according to the target level specification defined in Appendix~\ref{app:difficulty}. The agent's initial position and heading are then sampled from $\mathcal{P}$ within a 7.5\,m radius of $\mathbf{p}_{\text{amb}}$. Natural-language clues are automatically generated from the first-person observation sequence along the expert trajectory using a VLM with a task-specific prompt.

\paragraph{(3) Automatic episode execution and recording.} An built-in navigation system combined with our customized state machine drives the agent through the complete four-stage SAR workflow (Section~3.1), recording the full trajectory with observation snapshots (RGB, depth, semantic segmentation), waypoint sequences, and interaction annotations at trigger points. Unlike navigation-only pipelines that end at the goal coordinate, this pipeline records the complete interaction stream, including pre-interaction and post-interaction observation pairs as well as precise distance measurements to the interaction anchor. These annotations provide dense supervision for training approach behaviors specific to the $\delta_{\text{trig}} = 1.5\text{m}$ interaction protocol.

\paragraph{(4) Difficulty validation.} After episode generation, the pipeline applies the validation criteria defined in Appendix~\ref{app:difficulty}, including interaction requirements for higher levels and structural constraints for multi-floor settings. Episodes failing any criterion are rejected and re-sampled until a valid configuration is obtained.

The pipeline generates arbitrarily many validated episodes per environment and extends to any UE5 environment via the navigation mesh export plugin. Fine-tuning on automatically collected data yields consistent improvement across all adapted baselines (Section~\ref{app:results_ext}), confirming that the pipeline produces training-quality supervision without human involvement.

\subsection{Simulator Interface and Evaluation Protocol}
\label{app:simulator}

RescueBench is built on Unreal Engine 5.6 with UnrealCV~\cite{qiu2017unrealcv} and UnrealZoo~\cite{zhong2025unrealzoo}. The simulator matches the real-time control interval used by the environment. Camera resolution and field of view are not fixed benchmark constants; both can be modified through the API according to the requirements of a specific method. By contrast, the simulator itself does not impose a fixed policy inference frequency, since the effective action rate is determined by each model's runtime latency rather than by the environment.

Each agent receives synchronized RGB, depth, and semantic segmentation observations. Each method uses its official default input resolution; the per-method resolutions are listed in Table~\ref{tab:input_resolution}. Ground agents operate in a hybrid action space with two continuous action spaces: forward velocity, angular velocity, and three binary interaction actions (pick, drop, open door). Our framework also supports integrating aerial-ground cooperation settings. Specifically, aerial agents use continuous action space: forward velocity, angular velocity, altitude velocity.

\begin{table}[tbp!]
\centering
\caption{Per-method input resolution. Each method is configured with its official default resolution; no resolution-specific tuning is performed.}
\label{tab:input_resolution}
\begin{tabular}{lc}
\toprule
\textbf{Method} & \textbf{Input Resolution (H $\times$ W)} \\
\midrule
LLM-YOLO Planner & $640 \times 640$ \\
Uni-NaVid & $224 \times 224$ \\
ViNT & $640 \times 640$ \\
NoMaD & $640 \times 640$ \\
SG-Nav & \multicolumn{1}{c}{$640 \times 480$} \\
OmniNav & $640 \times 569$ \\
ROCKET-2 & $640 \times 360$ \\
SPF + CityWalker & $1920 \times 1080$ / $640 \times 360$ \\
\bottomrule
\end{tabular}
\end{table} 


Evaluation follows level-dependent time budgets rather than a single global timeout: episodes at $L_1$--$L_2$ are capped at 180\,s, $L_3$ at 240\,s, and $L_4$--$L_5$ at 300\,s. To ensure that evaluation isolates navigation capability rather than manipulation precision, all baseline methods except ROCKET-2 are equipped with automatic interaction triggers: once the agent reaches within $\delta_{\text{trig}} = 1.5$\,m of the target or handoff zone, the corresponding interaction is executed automatically. As a result, failures at S2 and S4 indicate that the agent did not navigate close enough to trigger the interaction, rather than an inability to manipulate the environment once there.

\section{Multi-Agent Extension}
\label{app:multi_agent}
This section documents preliminary multi-agent evaluation conducted to validate RescueBench's extensibility to heterogeneous agent configurations.

\subsection{Experimental Setup}
\label{app:multi_agent_setup}
\paragraph{Multi-Agent Configuration.} We manually create an aerial-ground configuration to validate the framework's extensibility to heterogeneous multi-agent settings. 

Specifically, we employ an aerial UAV agent running SPF~\cite{yin2024sg} conducts exploration from elevated vantage points for wider visual coverage, while a ground humanoid agent running CityWalker~\cite{liu2025citywalker} serves as the primary rescuer for Stages 2--4. Upon target detection by the UAV, the ground agent is directed to the target location to complete the rescue pipeline. Since the UAV's elevated viewpoint makes short-range target detection trivial, multi-agent evaluation is restricted to $L_3$--$L_5$.

The heterogeneous SPF + CityWalker setup uses a flight altitude of 3\,m for the UAV to provide broad overhead coverage during search. SPF and CityWalker are fine-tuned independently on single-agent data only; no multi-agent-specific joint training is introduced.

\subsection{Results and Analysis}
Table~\ref{tab:multi_agent_overall} and Table~\ref{tab:multi_agent_stage} report multi-agent results.

\begin{table}[tbp!]
\centering
\caption{Multi-agent (SPF + CityWalker) overall performance across difficulty levels L3--L5.}
\label{tab:multi_agent_overall}
\begin{tabular}{c|c|c|c}
\toprule
\textbf{Level} & \textbf{TCR (\%)} & \textbf{TS (0--100)} & \textbf{Avg Time (s)} \\
\midrule
L3 & 10.5 & 30.6 & 228.4 \\
L4 & 7.1 & 14.3 & 288.1 \\
L5 & 0.0 & 1.6 & 301.4 \\
\bottomrule
\end{tabular}
\end{table}

\begin{table}[tbp!]
\centering
\caption{Multi-agent (SPF + CityWalker) per-stage Task Score (/25) across difficulty levels.}
\label{tab:multi_agent_stage}
\begin{tabular}{c|c|c|c|c|c|c}
\toprule
\textbf{Level} & \textbf{S1 (Explore)} & \textbf{S2 (Rescue)} & \textbf{S3 (Return)} & \textbf{S4 (Handoff)} & \textbf{Avg Steps} & \textbf{FPS} \\
\midrule
L3 & 11.8 & 10.9 & 3.9 & 3.9 & 533.3 & 2.2 \\
L4 & 7.1 & 3.6 & 1.8 & 1.8 & 633.0 & 2.2 \\
L5 & 1.6 & 0.0 & 0.0 & 0.0 & 615.0 & 2.2 \\
\bottomrule
\end{tabular}
\end{table}

\textbf{Multi-agent (SPF + CityWalker)} demonstrates that aerial search is more efficient than a single ground agent for finding the target region, yet the bottleneck persists in ground-level open-world navigation under 3D structural complexity. At $L_3$ and $L_4$, aerial-ground cooperation achieves S1 scores second only to ROCKET-2 (11.8 and 7.1), but the subsequent stages reveal the same spatial memory deficit observed in single-agent baselines. Critically, this cooperation paradigm relies on sharing absolute GPS coordinates between UAV and ground agent, which fails in GPS-denied environments such as indoor structures, dense urban canyons, or subterranean settings common in real SAR operations. This finding reveals that future multi-agent frameworks should explore shared semantic representations and relative spatial reasoning that remain robust under partial observability, rather than depending on fragile absolute coordinate exchange.

\begin{figure}[tbp!]
    \centering
    \includegraphics[width=\linewidth]{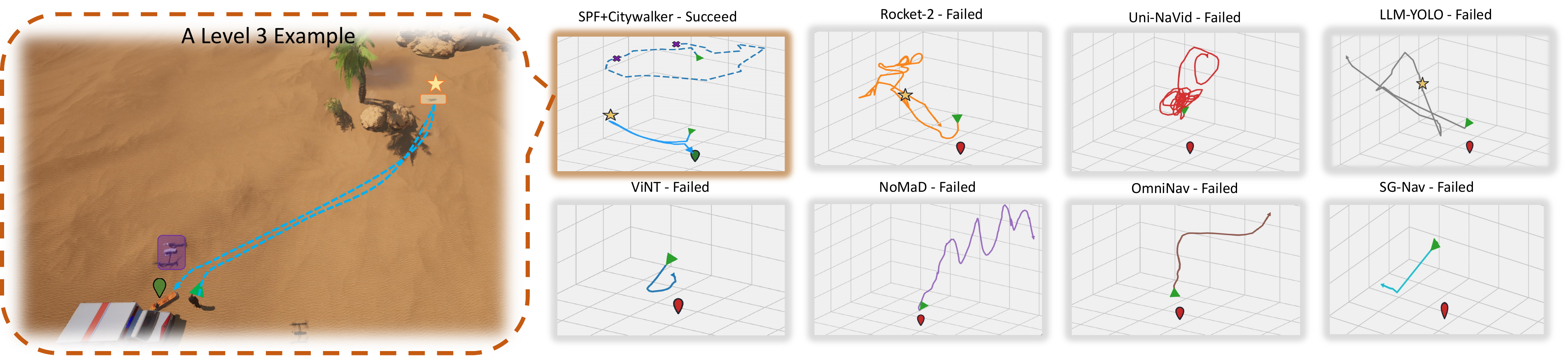}
    \caption{Trajectory visualization of a complete four-stage successful episode of SPF + CityWalker in DesertMap. The aerial UAV agent conducts progressive search, first locating the injured person and then the ambulance, transmitting GPS coordinates to the ground agent at each discovery (marked by purple crosses). The ground CityWalker agent navigates to the injured person for rescue (S2) and subsequently returns to the ambulance using the relayed location (S3).}
    \label{fig:multi_agent_traj}
\end{figure}

\section{Metric Details}
\label{app:metric_details}

\subsection{Human Similarity Metric}
\label{app:hs_detail}

The Human Similarity (HS) metric uses Dynamic Time Warping (DTW)~\cite{sakoe1978dynamic} to compare an agent's trajectory to the corresponding human expert demonstration on the same episode. Unlike pointwise metrics such as cosine similarity, which compare instantaneous movement directions at each step and require equal-length, temporally aligned sequences, DTW finds the optimal non-linear alignment between two sequences by warping the time axis. The DTW distance is computed as the minimum cumulative cost over all valid warping paths:

\[
\text{DTW}(\boldsymbol{\tau}^A, \boldsymbol{\tau}^E) = \min_{\pi} \sum_{(i,j) \in \pi} \|\mathbf{p}_i^A - \mathbf{p}_j^E\|_2,
\]

where $\boldsymbol{\tau}^A$ and $\boldsymbol{\tau}^E$ are the 2D waypoint sequences of the agent and expert respectively, and $\pi$ is a warping path satisfying monotonicity and boundary constraints. The distance is then mapped to $[0,1]$ via exponential normalization:

\[
\text{HS} = \exp\!\left(-\frac{\text{DTW}(\boldsymbol{\tau}^A,\; \boldsymbol{\tau}^E)}{\sigma}\right), \quad \sigma = \text{median}(\|\boldsymbol{\tau}^E\|_{\text{path}}).
\]

The normalization factor $\sigma$ is set to the median total path length of human demonstrations across all levels (approximately 42\,m), ensuring that HS values are comparable across episodes of varying spatial scales. An HS of 1.0 indicates identical trajectories; values near 0 indicate fundamentally different movement patterns.

\textbf{Diagnostic interpretation.} HS captures trajectory-level strategy rather than step-level precision. An agent following a rigid heuristic such as wall-following may achieve moderate TS but low HS, because its movement pattern is structurally dissimilar to human search behavior. Conversely, an agent that explores in human-like patterns (broad spatial coverage, systematic search) can achieve high HS even when TS is low due to imperfect termination. Together with TS and Avg Steps, HS enables three-way diagnosis: high HS + low TS → correct strategy, poor execution; low HS + high Avg Steps → inefficient strategy, wasteful movement; low HS + low Avg Steps → premature termination without meaningful exploration.





\end{document}